\documentclass[lettersize,journal]{IEEEtran}

\usepackage{amsmath,amsfonts}
\usepackage{algorithmic}
\usepackage{algorithm}
\usepackage{array}
\usepackage[caption=false,font=normalsize,labelfont=sf,textfont=sf]{subfig}
\usepackage{textcomp}
\usepackage{stfloats}
\usepackage{url}
\usepackage{verbatim}
\usepackage{graphicx}
\usepackage{cite}
\usepackage{caption}
\usepackage{booktabs}
\usepackage{xcolor, colortbl}
\usepackage[pagebackref,breaklinks,colorlinks,citecolor=cvprblue]{hyperref}

\definecolor{cvprblue}{rgb}{0.21,0.49,0.74}
\definecolor{Gray}{gray}{0.9}
\definecolor{Gray2}{gray}{0.95}
\definecolor{green}{RGB}{30,144,255}
\definecolor{customblue}{RGB}{34,113,179}
\definecolor{orangenum}{RGB}{255,120,0}

\newcommand\ie{\emph{i.e.}} 
\newcommand{\etal}{\textit{et al.}}
\newcommand\eg{\emph{e.g.}} 

\usepackage{enumitem}

\newcommand{\bluenum}[1]{\textcolor{customblue}{#1}}

\begin{document}

\title{Enhancing Vision-Language Models' Generalization via Diversity-Driven Novel Feature Synthesis}

        % <-this % stops a space


\author{Siyuan Yan, Cheng Luo, Zhen Yu, and Zongyuan Ge, \IEEEmembership{Senior Member, IEEE}
\thanks{S. Yan, C. Luo, Z. Yu, and Z. Ge is with the Monash University, Clayton, VIC. 3800 Australia, and also with Airdoc-Monash Research, Monash University, Clayton, VIC. 3800 Australia (E-mail: siyuan.yan@monash.edu, zongyuan.ge@monash.edu).}}

\markboth{Journal of \LaTeX\ Class Files,~Vol.~14, No.~8, August~2021}%
{Shell \MakeLowercase{\textit{et al.}}: A Sample Article Using IEEEtran.cls for IEEE Journals}

\IEEEpubid{0000--0000/00\$00.00~\copyright~2021 IEEE}

\maketitle

\IEEEpubidadjcol

\begin{abstract}

Vision-language foundation models like CLIP have shown impressive zero-shot generalization, but finetuning on downstream datasets can cause overfitting and loss of its generalization ability on unseen domains. Although collecting additional data from new domains of interest is possible, this method is often impractical due to the challenges in obtaining annotated data. To address this, we propose a plug-and-play feature synthesis method called LDFS (\textbf{L}anguage-Guided \textbf{D}iverse \textbf{F}eature \textbf{S}ynthesis) to synthesize new domain features and improve existing CLIP fine-tuning strategies. LDFS has three main contributions:  1) To synthesize novel domain features and promote diversity, we propose an instance-conditional feature augmentation strategy based on a text-guided feature augmentation loss. 2) To maintain feature quality after augmenting, we introduce a pairwise regularizer to preserve augmented feature coherence within the CLIP feature space. 3) We propose to use stochastic text feature augmentation to reduce the modality gap and further facilitate the process of text-guided feature synthesis.  Extensive experiments show LDFS's superiority in improving CLIP's generalization ability on unseen domains without collecting data from those domains. The code will be made publicly available.

\end{abstract}    


\section{Introduction}
\label{sec:intro}

\IEEEPARstart{I}{n} recent years, vision-language foundation models, such as CLIP (Contrastive Language-Image Pre-training) \cite{clip}, have achieved significant advancements in the field of computer vision. CLIP is trained to align the text and image semantic spaces from a text and image encoder respectively by learning from 400 million image-text pairs. The pre-trained model can achieve impressive zero-shot generalization ability by providing text prompts such as ``a photo of a [class]”.
\begin{figure*}[!t]
    \centering
   \includegraphics[ width=0.8\linewidth]{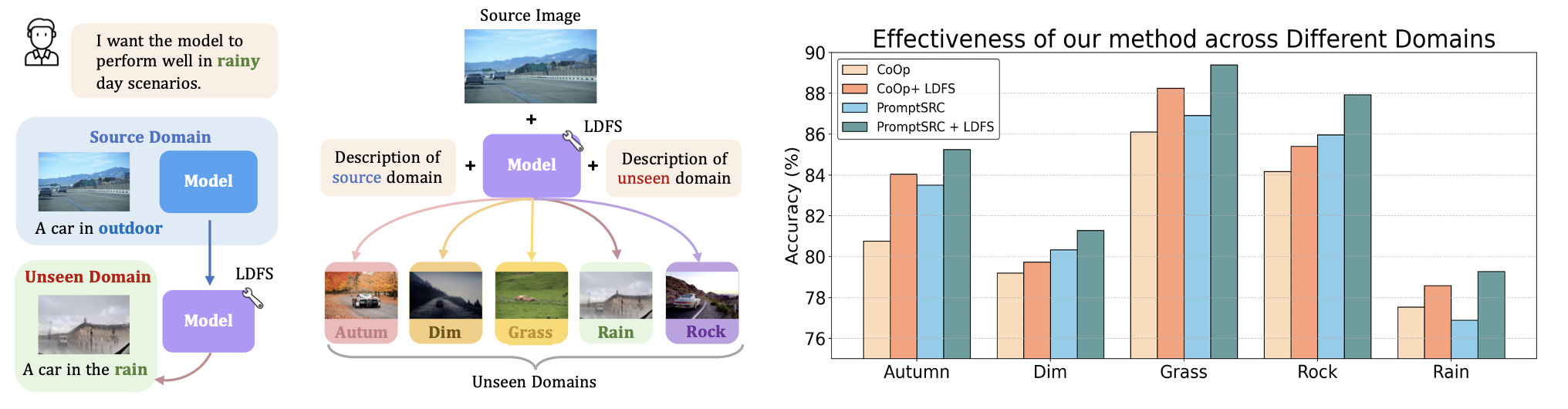} 
        % \caption{Illustration of the language-guided zero-shot domain adaptation.} 
    \caption{ Our LDFS facilitates model adaptation to unseen domains using language description only, negating the need for direct data access from those new domains. Additionally, LDFS acts as a plug-and-play feature augmentation method, enhancing the efficacy of CLIP-based finetuning approaches in unseen domains. }
    \label{fig1}
\end{figure*}

In order to further adapt CLIP to downstream tasks, some works propose to fine-tune CLIP via either linear probing or learning optimal soft prompts, such as CoOp \cite{coop}, CoCoOp \cite{cocoop}, MaPle \cite{maple}, and PromptSRC \cite{promptsrc}, in a process called prompt learning \cite{coop,p1,p2,p3} for downstream datasets. Although these fine-tuning methods can improve CLIP's performance on downstream datasets, a significant issue is that they can also cause the model to overfit to these datasets, resulting in even worse performance than the original zero-shot CLIP on some related datasets with domain-shift, limiting real-world applicability. One method is to collect additional data from the target domains of interest. However, this approach is often impractical due to the challenges in obtaining data from new or rare domains, further compounded by the costs of annotation for effective adaptation.

To address this issue, we aim to develop a feature augmentation method that can synthesize visual features from novel domains by leveraging the knowledge existing in pre-trained CLIP. Our motivation is based on the observation that the domain-shift direction in CLIP's text embedding can guide image feature augmentation in CLIP's image embedding, without requiring access to any images from the target domain \cite{clipstyler,lads,clipnada}. However, directly manipulating the text latent space of CLIP to synthesize either novel domain features (e.g., LADS \cite{lads}) or images (e.g., CLIPStyler \cite{clipstyler}) cannot effectively improve CLIP's performance on unseen domains. We attribute this to the tendency of these methods to synthesize visual features that lack diversity within the same class (e.g., LADS in Fig. \ref{fig4}) or generate low-quality images with artifacts (e.g., CLIPStyler in Fig. \ref{figstyler}).
\IEEEpubidadjcol
In this paper, we propose LDFS (\textbf{L}anguage-Guided \textbf{D}iverse \textbf{F}eature \textbf{S}ynthesis), which synthesizes diverse and high-quality novel domain visual features given source domain training data along with text descriptions of the training data and the target domain of interest only, as illustrated in the left of Fig. \ref{fig1}. We achieve high diversity and high-quality synthesized features using three approaches. 1)\textit{Promoting feature diversity with instance-conditional augmentation:} We notice that existing text-guided feature augmentation and styler transfer methods \cite{lads,clipnada,clipstyler} use a shared, global augmentation direction for all images in each category, leading to homogeneous synthesized features or images. We address this by introducing a novel adaptation loss that utilizes more nuanced, instance-conditional text descriptions instead of uniform descriptions. This enables our method to synthesize more diverse visual features. As illustrated in Fig. \ref{f2}, rather than applying a global direction from ``a photo of a horse'' to ``a sketch of a horse'' for all horse images, our method employs customized descriptions like ``a (sketch) photo of a horse with its head raised'' or ``a (sketch) photo of a horse grazing on a green field'' for individual images. This strategy ensures that the augmentation process is informed by the unique attributes of each image, enhancing feature diversity and capturing richer visual semantics. 2) \textit{Preserving feature coherence in adapted space:} We observe that the feature augmentation process could inadvertently drift synthesized features from the CLIP sphere, negatively impacting the synthesis quality and losing the class information. This occurs because the CLIP-based augmentation loss overlooks the preservation of pairwise cosine distances for adapted features. To mitigate it, we propose a regularizer to explicitly maintain the pairwise relationship for augmented features so that augmented features lie in the CLIP sphere, achieving better synthesis quality. 3) \textit{Reducing the cross-modality representation gap:} Despite CLIP's ability to represent text and image features in a joint space, there persists a modality representation gap, as documented in recent studies \cite{gap1,gap2}.  We notice that this modality gap weakens the integration of cross-modality features and thus hampers the adaptation process. To address it, we propose adjusting the text representation closer to its image counterpart by introducing controlled perturbations.

Overall, LDFS is a flexible two-stage framework. In the first stage, it transforms the training data into diverse target domain visual features guided by language description in the text space of CLIP. In the second stage, it finetunes CLIP on both original features and newly synthesized ones. LDFS can improve various CLIP finetuning strategies such as linear probing, CoOp \cite{coop}, CoCoOp \cite{cocoop}, MaPle \cite{maple}, and PromptSRC \cite{promptsrc}, and will boost their performances in unseen domains, as shown in the right of Fig \ref{fig1}. We validate its efficacy through extensive experiments on several challenging domain-shift benchmarks, including PACS \cite{pacs}, Office-Home \cite{oh}, DomainNet \cite{domainnet}, and NICO++ \cite{nico++}.

In summary, our main contributions are: (1) We identify homogeneous feature synthesis issues and address them through a novel instance-conditional adaptation strategy. (2) We recognize and mitigate the modality gap issue via a stochastic text augmentation method, facilitating more effective feature synthesis. (3) We maintain the adapted features within the CLIP sphere through a novel regularization loss, improving the quality of the augmented features. (4)  Quantitative and qualitative results demonstrate LDFS's superiority in achieving enhanced, diversified feature synthesis and significantly improving existing CLIP's fine-tuning methods on unseen domains.

\section{Related Work}
\label{sec:related work}

\noindent\textbf{Vision and Language Representation.}
Vision-language pretraining models \cite{clip,align,vl} establish their efficacy by learning visual and language modalities in a joint representation space. For instance, CLIP was trained on 400M text-image pairs, employing contrastive loss and drawing from a vast, web-scale dataset. The aligned representation learned by contrastive loss \cite{c_loss} provides CLIP with strong zero-shot capabilities. In the realm of downstream applications, prompt learning is a common strategy for adapting CLIP that learns task-specific knowledge through additional learnable tokens, thereby avoiding the expensive re-training of the pre-trained model. Representative methods such as CoOp \cite{coop} and CoCoOp \cite{cocoop} enhance CLIP's adaptability via learning continuous tokens and image-conditional continuous tokens, respectively. Furthering this trajectory, recent works such as MaPle \cite{maple} and PromptSRC \cite{promptsrc} leverage multi-modal prompt learning for better transfer learning. Nonetheless, existing finetuning methods are not tailored for the domain shift scenario. To bridge the gap, we propose to synthesize diverse target domain features to facilitate unseen domain adaptation. Our method is compatible with existing finetuning methods of CLIP, enhancing their efficacy under domain shift.\\
\noindent\textbf{Text-guided Image Synthesis.}
Text-guided image synthesis \cite{clipnada,clipstyler,styleclip,diffusionclip} aims to modify source images through text descriptions. For example, StyleGAN-NADA \cite{clipnada} achieves zero-shot image synthesis by leveraging text prompts, identifying the domain shift in the text space as a directional to guide image generation. Similarly, CLIPStyler \cite{clipstyler} finetunes a style transfer network utilizing CLIP embeddings for text-driven style transfer. While these works make progress in generative tasks, they are not designed for tackling domain shift problems and necessitate additional generative models. Moreover, their efficacy is frequently constrained by the quality of the generation process, which often introduce artifacts or wrong classes for downstream adaptation (see Fig. \ref{figstyler}). Similarly, LADS \cite{lads} and PODA are introduced, proposing to directly manipulate CLIP embeddings to synthesize features for model adaptation. Nevertheless, a common issue among all the above methods is their reliance on a global adaptation direction within the CLIP space, which limits the diversity of synthesized images or features. This sometimes leads to even worse performance than zero-shot CLIP (e.g., LADS compared to CLIP (ZS) in Table \ref{tab1} to \ref{tab4} and CLIPStyler compared to CLIP (ZS) in Table \ref{tab6}. In contrast to prior efforts, we introduce an instance-conditional local direction adaptation framework to promote both diverse and high-quality visual feature synthesis for effective downstream adaptation.\\
\noindent\textbf{Model Generalization.} Domain adaptation (DA) \cite{adv1,adv2,epvt} aims to bridge the distribution gap between the source and target domains so that the model can generalize in target domains. In unsupervised domain adaptation, this is typically achieved by leveraging additional unlabeled data from the target domain and utilizing techniques such as adversarial training \cite{adv1,adv2,adv3}, and moment matching \cite{moment1,moment2,moment3}, to minimize distribution disparities, all while reducing the necessity for costly annotations in the target domain. More recently, few-shot domain adaptation \cite{fda1,fda2} has been introduced, allowing models to adapt to a new domain with only a few examples, further relaxing the data requirement. Diverging from DA, our method does not require any data from target domains. Instead, it utilizes descriptive text about the source and target domains, guiding the adaptation process of source images via CLIP \cite{clip} models. It's noteworthy that, while domain generalization \cite{dg} also does not require target domain data, it typically requires training on multiple source domain data. Although some methods \cite{sdg1,sdg2,pldg} only require a single source domain data (called single domain generalization), they often lead to limited improvements in data with domain shift. This is in contrast to our method, which is trained with just a single source domain and significantly enhances CLIP's fintuning performance.

\section{Method}
 \begin{figure*}[!t]
   \begin{center}
   {\includegraphics[width=1\linewidth]{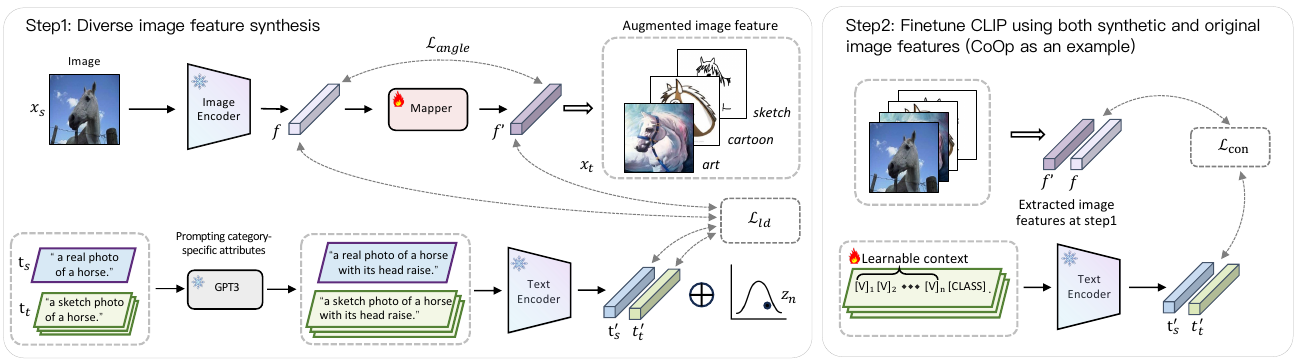}}
   \end{center}
%   \vspace{-10mm}

  \vspace{-2mm}
\caption{\small In Step 1 of our LDFS framework, a mapper network employs $\mathcal{L}_{ld}$ to translate source domain image features into diverse target domain features guided by a captioning model, ensuring instance-conditional text. Stochastic text feature augmentation is used to bridge the gap between image and text modalities, while a pairwise relation regularization loss $\mathcal{L}_{pair}$ aligns angles between original CLIP features and synthesized features, enhancing synthetic quality. In Step 2, CLIP is finetuned using both synthetic and original features for downstream adaptation through cross-entropy loss $\mathcal{L}_{con}.$
}
\label{method1}
\end{figure*}
In this section, we present our framework based on an instance-conditional local adaptation strategy, a pair-wise regularizer, and text feature augmentation, which synthesizes diverse and high-quality target domain features rather than using original features only. To formulate the task, consider a training set $\mathcal{D}_{train}=\lbrace (x_{s},y_{s})\rbrace_{s=1}^{S} $ where $x_{s}$ is an input image from source domain, $y_{s}$ is the corresponding ground truth, and S is the size of the source training dataset. The task also includes the description $t_{s}$ of the source training domain (e.g., ``a real photo of a \{\}'') plus the class name (\eg~horse), and a collection of $k$ descriptions $\lbrace t^{i}_{t}\rbrace_{i=1}^k$ of the $k$ unseen target domains $\lbrace \mathcal{D}^{i}_{unseen}\rbrace_{i=1}^{k}$ that cover the same classes as the source dataset. The goal is to adapt the model to work well on the $k$ unseen target domains during testing time.

Our two-stage method, shown in Fig.~\ref{method1}, begins with extracting source image features $f=I(x_{s})$ using CLIP's frozen image encoder. Then, we enrich both source $t_s$ and target text descriptions  $t_t$ with attributes by querying GPT3. The refined descriptions are encoded by CLIP's text encoder into $t'_s$ and $t'_t$. To bridge the modality gap, we apply stochastic, data-independent noise to span the space of text features. A mapper network then maps the source image features $f$ into diverse target features $f'$. The mapping process is guided by the domain gap direction between $t'_{s}$ and $t'_{t}$ using a local adaptation loss $\mathcal{L}_{ld}$. We also employ a sphere regularizer to constrain the synthetic features $f'$ within the CLIP sphere. In stage 2, we combine the source image features $f$ with synthetic features $f'$ and finetune them to enhance the model performance in unseen domains.

\subsection{Text-Guided Diverse Feature Synthesis}
\begin{figure}[!t]
  \centering
  \includegraphics[width=0.48\textwidth]{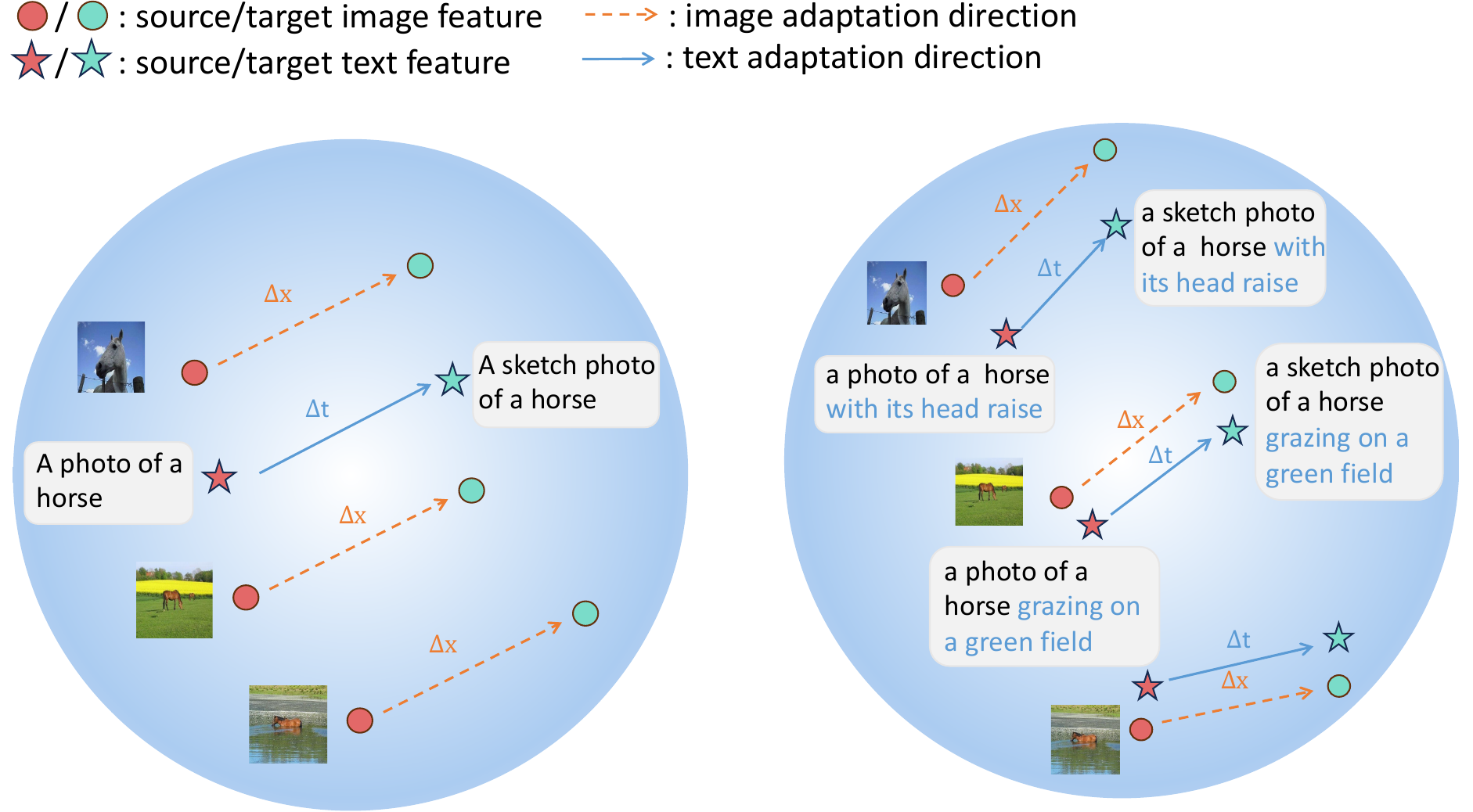}

  \begin{minipage}{0.24\textwidth}
    \centering
    \captionsetup{font=footnotesize}
    \vspace{3pt}
    \captionof*{figure}{(a) Global direction adaptation}
    \vspace{2pt}
  \end{minipage}%
  \begin{minipage}{0.24\textwidth}
    \centering
    \captionsetup{font=footnotesize}
    \vspace{3pt}
    \captionof*{figure}{(b) Local direction adaptation}
    \vspace{1pt}
  \end{minipage}
     \vspace{-2mm}

  \caption{Prior CLIP-based adaptation losses \cite{clipnada,lads,clipstyler} employ a consistent, global text description direction (i.e., $\Delta t$ in Fig. \ref{f2} (a)) for translating image features to the target domain. In contrast, our loss (Fig. \ref{f2} (b)) utilizes instance-specific local text description directions, enhancing the diversity of adapted image features.}
  \label{f2} 
\end{figure}

 \noindent\textbf{Instance-conditional local adaptation loss:}
In the generative model community, existing text-guided image synthesis methods use CLIP embeddings to guide additional generative models like StyleGAN to translate images through losses like global CLIP loss \cite{styleclip} or directional CLIP loss \cite{clipnada,lads}. However, existing CLIP losses utilize a fixed global direction for all images in a class, limiting the diversity of synthetic images, as illustrated in Fig. \ref{f2} (a). We propose utilizing instance-conditional direction to promote the diversity of synthetic features as it can construct customized text description directions for each image rather than a fixed shared text description direction, as illustrated in Fig. \ref{f2} (b). To create attributes-level text, we query GPT3 \cite{gpt3} to describe each source image with a semantically relevant attribute for each class through the following question:
\begin{enumerate}[label=\textbf{Q:},leftmargin=*,labelsep=1em]
  \item What are useful visual attributes for distinguishing a \{category\} in a photo?
\end{enumerate}
Notice that the generated attributes does not necessary describe the corresponding image, but it is used to improve the synthetic feature diversity.
% (e.g. ``a photo of a'') to obtain image-specific descriptions. The extracted attributes are consistent between source and target images, only differing in the ``domain'' description. For a detailed example, refer to Fig. \ref{method1}, with additional information in Appendix 1.3 explaining the image-specific description construction.

After obtaining potential visual attributes, we select relevant attributes by ranking their cosine similarity scores to image features within the CLIP joint space. Subsequently, we train a mapper network with our proposed instance-conditional local adaptation loss $\mathcal{L}_{ld}$ where the optimization objective is defined as aligning the normalized difference between each source image feature $I(x_{s})$ and the adapted feature $F_{m}^{k} ( I(x_{s}))$ with the normalized difference between the image-specific target domain description feature $t'_{t}$ and the corresponding source domain description feature $t'_{s}$ calculated by the text encoder $T$: 
\begin{equation}
 \vspace{2mm}
\label{eq1}
 \footnotesize
\begin{aligned}
&\Delta T_{s} =Norm(t'_{t})-Norm(t'_{s}) , \\
&\Delta I_{s} = Norm( F_{m}^{k} ( I(x_{s})))-Norm(I(x_{s})), \\
&\mathcal{L}_{ld} = \mathbb{E}_{x_{s} \in \mathcal{D}_{train} } \sum_{s=1}^S(1 - \frac{\Delta I_{s} \cdot \Delta T_{s}}{|\Delta I_{s}||\Delta T_{s}|}).
\end{aligned}
\end{equation}
where $F_{m}^{k}$ is the mapper for the $k$th domain and $Norm$ is the $L_{2}$ normalization.

Following the practice of related works \cite{clipnada,lads}, we also employ a class consistency loss $\mathcal{L}_{cc}$ to preserve the class information during the augmentation process:
 \begin{equation}
\label{eq2}
 \footnotesize
\mathcal{L}_{cc} = \mathcal{L}_{\text{CE}}\left( \text{Softmax}\left( F_{m}^{k}\left( I\left( x_{s} \right) \right) \mathbin{\cdot} T\left( y_{s} \right) \right), y_{s} \right)
\end{equation}
where $\mathcal{L}_{\text{CE}}$ is the cross-entropy loss.
% \begin{figure}[!t]
%   \centering
%   \includegraphics[width=0.58\textwidth]{Figures/method2.pdf}

%   \begin{minipage}{0.44\textwidth}
%     \centering
%     \captionsetup{font=footnotesize}
%     \vspace{3pt}
%     \captionof*{figure}{(a) Global direction adaptation}
%     \vspace{2pt}
%   \end{minipage}%
%   \hspace{6mm}
%   \begin{minipage}{0.44\textwidth}
%     \centering
%     \captionsetup{font=footnotesize}
%     \vspace{3pt}
%     \captionof*{figure}{(b) Local direction adaptation}
%     \vspace{1pt}
%   \end{minipage}
%      \vspace{-7mm}

%   \caption{Prior CLIP-based adaptation losses \cite{clipnada,lads,clipstyler} employ a consistent, global text description direction (i.e., $\Delta t$ in Fig. \ref{f2} (a)) for translating image features to the target domain. In contrast, our loss (Fig. \ref{f2} (b)) utilizes instance-specific local text description directions, enhancing the diversity of adapted image features.}
%   \label{f2} 
% \end{figure}

 \noindent\textbf{Stochastic text feature augmentation:}
The effectiveness of CLIP-based adaptation loss relies on the assumption that image-text features are well aligned in the joint embedding of CLIP. However, Liang \etal \cite{gap1} demonstrated that there exists a modality gap in the CLIP embedding that can separate image and text features, impacting cross-modality transferability. We discern this modality gap also impacts adaptation in text-guided image synthesis. Separately, \cite{gap3} introduced perturbing image features to span the space of domains. Inspired by it, we propose stochastically perturb text features $t'_s$ and $t'_t$ to span the text representation space, bringing text features closer to image features before optimizing the adaptation loss $\mathcal{L}_{ld}$ in Eq. \ref{eq1}. Specifically, $t'_s$ and $t'_t$ are augmented by adding Gaussian noise $z$ sampled from a standard normal distribution and scaled by a hyperparameter $\beta$, such that $t'_s = t'_s + \gamma \cdot z$ and $t'_t = t'_t + \gamma \cdot z$. The perturbed text features are then renormalized. We show this stochastic augmentation can close the modality gap and improve image feature augmentation in section \ref{sec_ab}.

\noindent\textbf{Pairwise relation loss:} Similar to other CLIP-based adaptation losses \cite{lads,clipnada}, our $\mathcal{L}_{ld}$ loss in Eq.\ref{eq1} aims to adapt the image features towards the direction of text descriptions. However, this may make adapted features outside the CLIP sphere, rendering the cosine distance-based alignment $\frac{\Delta I_{s} \cdot \Delta T_{s}}{|\Delta I_{s}||\Delta T_{s}|}$ less effective. Intuitively, the CLIP space is like a unit sphere where cosine alignment works well for features that lie in the sphere, such as original CLIP features $f$. But with $\Delta I$ and $\Delta T$ in Eq.\ref{eq1}, adapted features may shift off this sphere since cosine distance is not preserved during the adaptation, leading to less effective alignment when using $\mathcal{L}_{ld}$. To keep adapted features on the CLIP sphere, we employ a pairwise loss to maintain relative relationships between adapted and source image features. This constraint acts as a regularizer to preserve the cosine distance relationships of the original CLIP features, improving synthesis quality, as defined as:
\begin{equation}
\label{eq3}
 \footnotesize
\mathcal{L}_{\text{pair}} = \sum_{{(s,w) \in S^2}} \left( \frac{\langle f_s, f_w \rangle}{|f_s| |f_w|} - \frac{\langle f'_s, f'_w \rangle}{| f'_s | | f'_w |} \right)^2
\end{equation}
where $f$ and $f'$ are the source and adapted image features. $s$ and $w$ are any two pair-wise point in the image feature $S^{2}$.

\noindent\textbf{Total loss in the synthetic stage:}  At this stage, our objective function consists of three parts: the instance-conditional local adaptation loss $\mathcal{L}_{ld}$, which facilitates diverse feature synthesis for unseen domains; the class consistency loss $\mathcal{L}_{cc}$, ensuring class information preservation during adaptation; and the pairwise relation regularizer $\mathcal{L}_{pair}$, responsible for maintaining consistent relationships between the adapted and source image features. Our total loss for the synthetic stage is then expressed as:
\begin{equation}
\label{eq4}
\mathcal{L}_{synthetic}=\mathcal{L}_{ld}+ \alpha\cdot\mathcal{L}_{cc}+ \beta \cdot \mathcal{L}_{pair}
\end{equation}
where $\alpha$ and $\beta$ are weighting hyperparameter.
\subsection{Finetuning CLIP with Synthesized Features}
After the training of the mapper network in stage 1, we obtain synthetic features from $k$ domains of interest, such as 
\begin{equation}
f'_s = \text{Concat}\left( F^1_m(I(x_s)), F^2_m(I(x_s)), \ldots, F^k_m(I(x_s)) \right)
\label{eq5}
\end{equation}
We then finetune CLIP using strategies like CoOp \cite{coop}, CoCoOp \cite{cocoop}, MaPle \cite{maple}, and PromptSRC \cite{promptsrc} by combining both original features $f_s$ and synthesized features $f'_s$. As the synthetic features have covered the distribution of unseen domains, their integration during finetuning can enhance the model's efficacy in previously unseen domains. The inference is the same as CLIP except using augmented image features rather than original features.

\section{Experiments}
\subsection{Experimental Setup}

\begin{table}[t!]
   \caption{\small Hyperparameters for training on the four datasets. LR1 and WD1 are the learning rate and weight decay in stage 1. LR2 and WD2 are the learning rate and weight decay in stage 2. \(\alpha\) and \(\beta\) are weighting hyperparameters for the total loss in stage 1. \(\lambda\) is the noise level for stochastic text feature augmentation.}

   \centering
   \small
   \scalebox{0.91}{
    \setlength{\tabcolsep}{1.6mm}{
    \begin{tabular}{cccccccc}
    \toprule[1pt]
    Datasets & LR1 & WD1 & LR2 & WD2 & \(\alpha\) & \(\beta\) & \(\lambda\) \\
    \midrule
    PACS \cite{pacs}       & 1e-3 & 5e-2 & 2e-3 & 5e-4 & 0.5  & 0.5  & 0.1  \\
    OfficeHome \cite{oh} & 1e-2 & 5e-2 & 2e-3 & 1e-4 & 0.5  & 1    & 0.01 \\
    DomainNet \cite{domainnet}  & 1e-2 & 5e-2 & 2e-3 & 1e-4 & 0.5  & 1    & 0.01 \\
    NICO++ \cite{nico++}     & 1e-2 & 5e-2 & 2e-3 & 1e-4 & 0.5  & 0.75 & 0.08 \\
    \bottomrule[1pt]
    \end{tabular}
    }
   }
\label{sup_tab1}
\end{table}
\begin{table}[t!]

   \caption{\small Top 1 accuracy \% on PACS (ViT-B/16). }
     % \vspace{-2mm}
   \centering
   \small
   \scalebox{0.73}{
    \setlength{\tabcolsep}{1.6mm}{
\begin{tabular}{lllll}
  \toprule[1pt]
Method & Art            & Cartoon       & Sketch        & Avg.       \\ \hline
CLIP (ZS) \cite{clip}                           & 97.23          & 99.10          & 87.81         & 94.77         \\
CLIP (LP)\cite{clip}                           & 86.62          & 90.98         & 83.92         &  87.17             \\
DPL \cite{dpl}                                 & 96.11           & 97.92          & 89.74         & 94.59         \\
LADS \cite{lads}                               & 94.38          & 97.01         & 88.22         & 93.20          \\
CoOp \cite{coop}                               & 95.41           & 97.70          & 90.67          & 94.59         \\
CoCoOp \cite{cocoop}                             & 96.58           & 98.72          & 90.96         & 95.42         \\
MaPle \cite{maple}                               & 96.95          & 98.99         & 88.79         & 94.91         \\
PromptSRC \cite{promptsrc}                          & 97.84          & 99.02         & 93.49         & 95.78         \\  
% \hline
\rowcolor{Gray2}
LP + LDFS  & 96.22 (\bluenum{+1.84})  & 98.85 (\bluenum{+1.84}) & 88.42 (\bluenum{+0.20}) & 94.50 (\bluenum{+1.29}) \\  
\rowcolor{Gray2}
CoOp + LDFS  & 97.61 (\bluenum{+2.20})  & 99.12 (\bluenum{+1.42}) & 92.78 (\bluenum{+2.11}) & 96.51 (\bluenum{+1.92}) \\                        
\rowcolor{Gray2}
CoCoOp + LDFS                       & 97.84 (\bluenum{+1.24})  & 99.19 (\bluenum{+0.49}) & 93.62 (\bluenum{+2.66}) & 96.88 (\bluenum{+1.46}) \\
\rowcolor{Gray2}
MaPLe + LDFS                        & 97.74  (\bluenum{+0.79}) & \textbf{99.41} (\bluenum{+0.42})  & 91.15 (\bluenum{+2.36}) & 96.19 (\bluenum{+1.19}) \\
\rowcolor{Gray2}
PromptSRC + LDFS                    & \textbf{99.12} (\bluenum{+1.28})  & 99.23 (\bluenum{+0.21}) & \textbf{95.04} (\bluenum{+1.55}) & \textbf{97.80} (\bluenum{+2.02})  \\ 
 \midrule[1pt]
\end{tabular}}}
\label{tab1}
\end{table}

\noindent\textbf{Datasets.}
We evaluate LDFS on four benchmark datasets to assess its generalization capabilities in unseen domains. \textbf{PACS} \cite{pacs} contains 9,991 images across 4 domains (Photos, Art, Cartoon, and Sketch) spanning 7 classes. \textbf{OfficeHome} \cite{oh} offers 15,588 images across 4 domains (Art, Clipart, Product, and Real) categorized into 65 classes. \textbf{DomainNet} \cite{domainnet} includes 586,575 images from 6 domains (Art, Clipart, Product, and Real) encompassing 345 classes, while \textbf{NICO++} \cite{nico++} contains 89,232 images over 6 domains (Outdoor, Autumn, Dim, Grass, Rock, Water) and 7 classes. We train models on the source domain (i.e., Photos for PACS, Real for OfficeHome and DomainNet, and Outdoor for NICO++) and evaluate performance on the remaining target domains.

\noindent\textbf{Baselines.} We compare three groups of methods: \textbf{a) Zero-shot methods}. CLIP (ZS) \cite{clip}, a zero-shot CLIP model with a text prompt such as ``A photo of a [CLS]''. PromptStyler \cite{promptstyler}, which simulates domain-shift by synthesizing diverse style word vectors via prompt learning. \textbf{b) CLIP's finetuning strategies}, the main baselines we aim to compare or improve, such as CLIP (LP) \cite{clip}, a linear classifier built upon the CLIP image embedding, and state-of-the-art prompt learning strategies for CLIP, namely CoOp \cite{coop}, CoCoOp \cite{cocoop}, MaPle \cite{maple}, PromptSRC \cite{promptsrc}, and DPL \cite{dpl}, a domain-based prompt learning method. \textbf{c) Text-guided image synthesis approaches}, the main baselines we compared. The most comparable baselines are the recently proposed LADS \cite{lads} and PODA \cite{poda}, as they also synthesize image features without employing an additional generative model. Nevertheless, we also compare a state-of-the-art image generation method, called CLIPStyler \cite{clipstyler}, focusing on style transfer for real image synthesis using an additional generative model, with finetuning using both original and synthesized images. It is noteworthy that we do not compare domain generalization methods trained on multiple source domains or some recent domain adaptation methods requiring target domain data (AD-CLIP \cite{ad-clip}, DAPL \cite{dapl}, CDTrans \cite{cdtrans}), as in some similar works \cite{poda,lads}, since they are unfair comparison. We also exclude comparison to recent muti-scale style-based prompt learning method Stylelip \cite{stylelip} as their code is unavailable, but given the consistent improvements of our method on different prompt learning strategies, LDFS should also be effective at improving Stylip.\\
\noindent\textbf{Implementation details.}
For PACS, OfficeHome, and DomainNet, we employ a ViT-B/16 based CLIP \cite{clip} model as our pre-trained vision-language model, as well as for other baselines. For NICO++, we use a ResNet50-based CLIP model. We build a 2-layer MLP as our mapping network and employ the ClipCap \cite{clipcap} as the captioning model. We report results averaged over 3 runs throughout all experiments. After conducting a grid search on crucial hyperparameters, such as learning rate and weight decay, the optimal configurations are selected based on in-domain validation accuracy for all methods. For prompt learning-based baselines and our methods, we follow the few-shot evaluation protocol in \cite{coop,clip}, using 16 images per class for training throughout the two stages and evaluating on complete test sets. However, we employ full-shot training in OfficeHome to evaluate our method's ability under the full-shot setting. All experiments are carried out on two NVIDIA GeForce RTX3090. The hyper-parameters are displayed in Table \ref{sup_tab1}. A for training parameters and GFLOPs, the training parameters for feature augmentation are 0.39M x the number of target domains, which is a very small portion of the parameter of ViT-b/16 based CLIP (91.16M+ 86.29M). GFLOPs (inference) is 187.4, which is comparable its baseline CoOp of 187.2.
\subsection{Evaluation}
\begin{table}[t!]
   \caption{\small Top 1 accuracy \% on OfficeHome (ViT-B/16). }
   % \vspace{-3mm}
   \centering
   \small
   \scalebox{0.73}{
    \setlength{\tabcolsep}{1.6mm}{
\begin{tabular}{lllll}
\toprule[1pt]
Method            &  Clipart & Product & Art & Avg.      \\\hline
CLIP (ZS) \cite{clip}         & 65.81                     & 88.93                     & 83.07                  & 79.27         \\
CLIP (LP) \cite{clip}         & 65.50                      & 88.51                     & 78.12                  & 77.38         \\
DPL \cite{dpl}              & 70.12                     & 90.16                     & 79.51                  &    79.90            \\
LADS \cite{lads}             & 69.53                     & 91.54                     & 81.25                  & 81.11        \\
CoOp \cite{coop}              & 70.41                      & 90.32                      & 80.80                   & 80.51         \\
CoCoOp \cite{cocoop}           & 70.02                        & 90.33                      & 80.78                   & 80.38         \\
MaPLe \cite{maple}            & 70.31                     & 92.23                     & 82.34                  & 81.63         \\
PromptSRC \cite{promptsrc}         & 72.95                     & 90.94                     & 83.41                  & 82.43         \\
\rowcolor{Gray2}
LP + LDFS  & 71.95 (\bluenum{+2.42})  & 91.21 (\textcolor{orangenum}{-0.33}) & 83.24 (\bluenum{+1.99}) & 82.13 (\bluenum{+1.02}) \\  
\rowcolor{Gray2}
CoOp + LDFS       & 71.80 (\bluenum{+1.39})     & 92.68 (\bluenum{+2.36})   & 84.39 (\bluenum{+3.59})& 82.96 (\bluenum{+2.45})   \\ \rowcolor{Gray2}
CoCoOp + LDFS     & 70.49 (\bluenum{+0.47})   & 92.19 (\bluenum{+1.86})   & 82.17 (\bluenum{+1.39})& 81.62 (\bluenum{+1.24}) \\ \rowcolor{Gray2}
MaPLe + LDFS      & 72.61 (\bluenum{+2.30})    & \textbf{94.25} (\bluenum{+2.02})   & 83.86 (\bluenum{+1.52})& 83.96 (\bluenum{+2.33}) \\ \rowcolor{Gray2}
PromptSRC +  LDFS & \textbf{74.30} (\bluenum{+1.35})    & 93.02 (\bluenum{+2.08})   & \textbf{85.31} (\bluenum{+1.90}) & \textbf{84.21} (\bluenum{+1.78}) \\
 \midrule[1pt]
\end{tabular}}}
\label{tab2}
\end{table}

\noindent\textbf{Main results.} First, we benchmark various SOTA algorithms and our LDFS on PACS, OfficeHome, DomainNet, and NICO++. The results are reported in Table \ref{tab1}, \ref{tab2}, \ref{tab3}, and \ref{tab4}. The main observations from existing algorithms include: 1) We notice linear probing, \ie~CLIP (LP), can hurt the generalization ability of the pre-trained CLIP model, as demonstrated by its subpar performance compared to zero-shot CLIP, \ie~CLIP (ZS) across all four datasets. 2) Existing CLIP-based finetuning methods, along with text-guided feature synthesis methods such as DPL, LADS, and PODA, are not always effective for target domain adaptation. For instance, both DPL and LADS degrade the average performance of zero-shot CLIP in PACS and NICO++, as shown in Table \ref{tab1} and \ref{tab4}. 

\begin{table*}[t!]
   \caption{Top 1 accuracy \% on DomainNet (ViT-B/16).}
   \centering
   \small
   \scalebox{1}{
    \setlength{\tabcolsep}{1.6mm}{
\begin{tabular}{lllllll}
\toprule[1pt]
Method         & Clipart       & Painting      & Sketch        & Infograph     & Quickdraw     & Avg.      \\\hline
CLIP (ZS) \cite{clip}      & 69.91         & 64.76         & 63.12         & 46.80          & 13.69         & 51.66         \\
CLIP (LP) \cite{clip}      & 70.12         & 59.82         & 62.42         & 48.17         & 14.17         & 50.94         \\
DPL  \cite{dpl}          & 70.16         & 67.91         & 64.89         & 50.22         & 12.01         & 53.04         \\
LADS \cite{lads}          & 72.46         & 67.16         & 64.04         & 50.70          & 14.16         & 53.70          \\
CoOp     \cite{coop}      & 71.60          & 66.86         & 63.85         & 51.77         & 11.91         & 53.20          \\
PromptSRC \cite{promptsrc}      & 72.63         & 66.67         & 65.20          & 52.89         & 14.01         & 54.28         \\  
\rowcolor{Gray2}
LP+ LDFS  & 72.49 (\bluenum{+0.03})  & 67.12 (\textcolor{orangenum}{-0.03}) & 64.32 (\bluenum{+0.28}) & 52.90 (\bluenum{+2.20})&14.57 (\bluenum{+0.41})&54.28(\bluenum{+0.58}) \\  
\rowcolor{Gray2}
CoOp + LDFS      & 73.54 (\bluenum{+1.94}) & \textbf{69.37} (\bluenum{+2.51}) & 65.81 (\bluenum{+1.96}) & 53.16 (\bluenum{+1.39}) & 16.08 (\bluenum{+4.17}) & 55.60 (\bluenum{+2.40}) \\ 
\rowcolor{Gray2}
PromptSRC + LDFS & \textbf{74.95} (\bluenum{+2.32}) & 69.03 (\bluenum{+2.36}) & \textbf{66.52} (\bluenum{+1.32}) & \textbf{54.32} (\bluenum{+1.43}) & \textbf{18.40} (\bluenum{+3.39})  & \textbf{56.64} (\bluenum{+2.36}) \\
 \midrule[1pt]
\end{tabular}}}
\label{tab3}
\end{table*}

\begin{table*}[t!]
   \caption{Top 1 accuracy \% on NICO++ (ResNet50).}
   \centering
   \small
   \scalebox{1}{
    \setlength{\tabcolsep}{1.6mm}{
\begin{tabular}{lllllll}
\toprule[1pt]
Method           & Autumn        & Dim           & Grass         & Rock          & Water         & Avg.      \\ \hline
CLIP (ZS) \cite{clip}        & 82.83         & 78.36         & 86.72         & 83.31         & 73.50          & 80.94         \\
CLIP (LP) \cite{clip}         & 68.20          & 71.96         & 77.82         & 78.09         & 73.55         & 72.35         \\
DPL \cite{dpl}              & 79.25         & 79.40          & 86.30          & 82.79         & 76.14         & 80.78         \\
LADS \cite{lads}             & 79.14         & 78.01         & 84.16         & 81.66         & 77.45         & 80.08         \\
PODA \cite{poda}                 & 78.73      &80.09   & 85.39         & 80.20         & 77.04         & 80.29         \\
CoOp \cite{coop}             & 80.76         & 79.20          & 86.10          & 84.17         & \textbf{79.53}         & 81.95         \\
MaPle \cite{maple}            & 82.36         & 77.46         & 85.32         & 85.01         & 77.19         & 81.47         \\
PromptSRC \cite{promptsrc}        & 83.50          & 80.34         & 86.91         & 85.96         & 76.89         & 82.92         \\  
\rowcolor{Gray2}
LP + LDFS  & 81.96 (\bluenum{+2.82})  & 78.29 (\bluenum{+0.28}) & 86.03 (\bluenum{+1.87}) & 82.93 (\bluenum{+1.27}) &77.06 (\textcolor{orangenum}{-0.39})&81.25(\bluenum{+1.17})\\  
\rowcolor{Gray2}
CoOp + LDFS         & 84.04 (\bluenum{+3.28})  & 79.73 (\bluenum{+0.53})  & 88.25 (\bluenum{+2.15}) & 85.40 (\bluenum{+1.23})   & 78.58 (\textcolor{orangenum}{-0.95})  & 83.20 (\bluenum{+1.25})   \\ \rowcolor{Gray2}
MaPle + LDFS      & 84.37 (\bluenum{+2.01})  & 79.94 (\bluenum{+2.48}) & \textbf{89.90} (\bluenum{+4.58})  & 86.14 (\bluenum{+1.13}) & 77.82 (\bluenum{+0.63})  & 83.63 (\bluenum{+2.16})  \\ \rowcolor{Gray2}
PromptSRC + LDFS & \textbf{85.25} (\bluenum{+1.75}) & \textbf{81.28} (\bluenum{+0.94}) & 89.40 (\bluenum{+2.49})  & \textbf{87.93} (\bluenum{+1.97}) & 79.27 (\bluenum{+2.38}) & \textbf{84.63} (\bluenum{+1.71}) \\
 \midrule[1pt]
\end{tabular}}}
\label{tab4}
\end{table*}

\begin{table}[t!]

   \caption{\small Comparison (\%) with the generative models. }
   \centering
   \small
   \scalebox{0.98}{
    \setlength{\tabcolsep}{1.6mm}{
\begin{tabular}{lllll}
  \toprule[1pt]
Method & Art            & Cartoon       & Sketch        & Avg.       \\ \hline
CLIP (ZS) \cite{clip}                           & 97.23          & 99.10          & 87.81         & 94.77         \\
% \hline
% \rowcolor{Gray2}
CoOp + CLIPStyler \cite{clipstyler}  & 96.33 & 98.19 & 89.06  & 94.53 \\        
CoOp + CLIPStyler\dag \cite{clipstyler}  & 95.38 & 97.52 & 89.15  & 94.02 \\    
% \rowcolor{Gray2}
CoOp + LDFS                       & \textbf{97.84}   & \textbf{99.19}  & \textbf{93.62}  & \textbf{96.88}  \\
 \midrule[1pt]
\end{tabular}}}
\label{tab5}
\end{table}

\begin{table}[t!]

   \caption{\small Synthetic feature quality (\%) on PACS . }
   \centering
   \small
   \scalebox{1}{
    \setlength{\tabcolsep}{3.1mm}{
\begin{tabular}{lccc}
  \toprule[1pt]
 & LADS           & CLIPStyler       & LDFS           \\ \hline
DA score                           & 98.70        & 79.41          & \textbf{99.15}              \\
% \hline
% \rowcolor{Gray2}
CC score   & 87.22 & 73.72 & \textbf{92.43}  \\                        
% \rowcolor{Gray2}
DS score                   & 66.71 & 95.87& \textbf{97.33}   \\
 \midrule[1pt]
\end{tabular}}}
\label{tab6}
\end{table}

Further, adding our LDFS to linear probing or current prompt learning techniques like CoOp, CoCoOp, MaPle, and PromptSRC results in significant performance improvements, as highlighted by blue numbers in Table \ref{tab1} to \ref{tab4}. Specifically, our LDFS combined with PromptSRC outperforms all baselines on average accuracy across all datasets while also achieving the best accuracy on most target domains. Also, it improves PromptSRC alone by 2.02\%, 1.78\%, 2.36\%, and 1.71\% respectively on four datasets. Additionally, LDFS significantly improves the efficiency of existing prompt learning techniques for certain domains. For example, CoOp + LDFS achieves a 3.11\% improvement over CoOp in the Art domain of OfficeHome and a 4.17\% improvement over CoOp in the Quickdraw domain of DomainNet. Similarly, MaPle + LDFS improves over MaPle by 4.58\% on the Grass domain of NICO++. In the linear probing setting, LDFS (LP+LDFS) outperforms LADS and PODA on average accuracy on NICO++. Notably, since PODA is specifically designed for CNN architectures, no PODA results are available beyond Table \ref{tab4}.\\
\begin{figure*}[t!]
    \centering
    \setlength{\tabcolsep}{5pt}  % Adjust this value to get desired spacing
    \renewcommand{\arraystretch}{0.9}
    \begin{tabular}{c@{\hspace{10pt}}c}
        \includegraphics[width=0.476\linewidth]{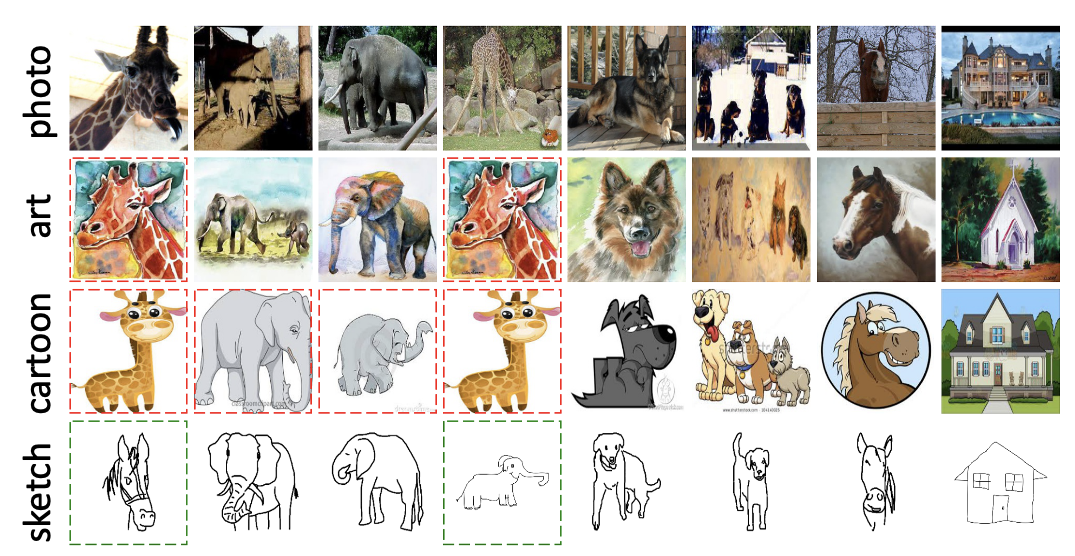} &
        \includegraphics[width=0.476\linewidth]{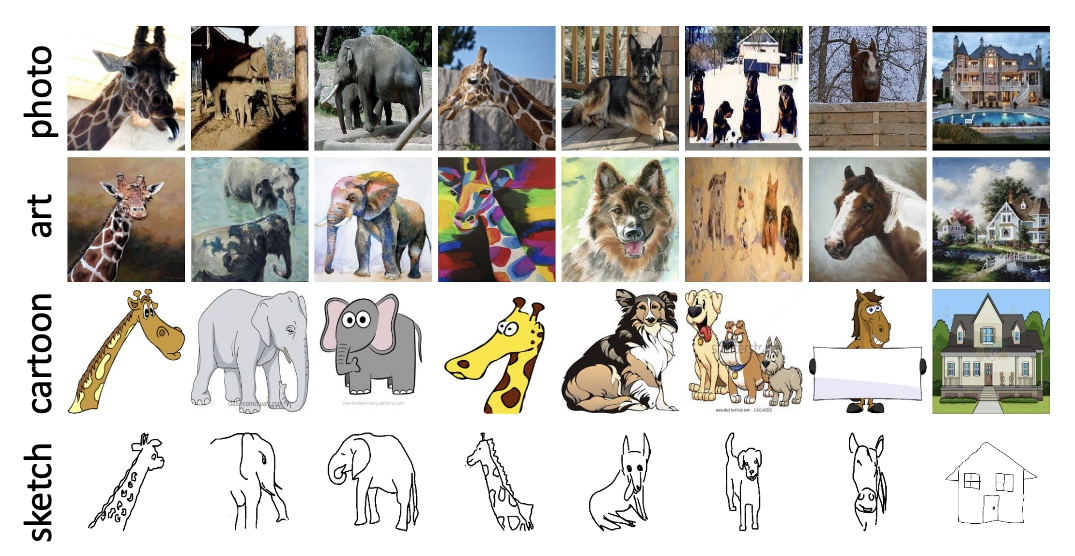} \\
        \footnotesize{(a) Synthesized feature nearest neighbors using LADS \cite{lads}.} & \footnotesize{(b) Synthesized feature nearest neighbors using LDFS (ours).} \\
    \end{tabular}
\caption{\small \textbf{Comparison of nearest neighbor images of synthesized target domain features between LADS and our LDFS on the PACS dataset: photo → art, cartoon, and sketch.} The first row displays randomly sampled images from the source domain (photo). Subsequent rows showcase the nearest neighbors of synthetic features from both LADS and LDFS. Red boxes highlight LADS' issues with homogeneous synthetic features, while green boxes indicate LADS' inability to preserve class information during adaptation.}
    \label{fig4}
\end{figure*}
\begin{figure}[!t]
    \centering
        \includegraphics[ width=0.95\linewidth]{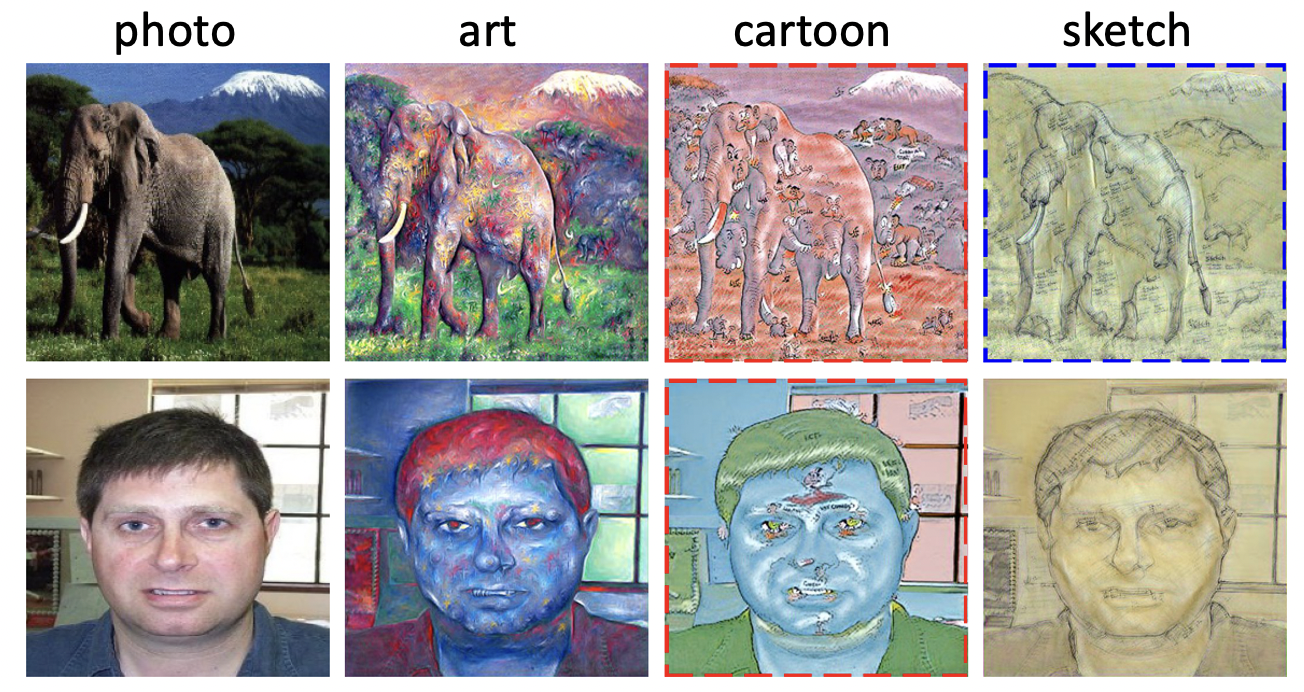} 
    \caption{\small \textbf{Style transfer photo images into art, cartoon, and sketch domains via CLIPStyler \cite{clipstyler}.} Unexpected artifacts (red box) and the model's inability to synthesize target objects (blue box) hinder its efficacy for downstream adaptation.}
    \label{figstyler} 
\end{figure}
\begin{figure}[t]
    \centering
    \setlength{\abovecaptionskip}{0.cm}
    \setlength{\belowcaptionskip}{-0.cm}
    \setlength{\tabcolsep}{1pt}  % <-- Adjust this value to get desired spacing
    \begin{tabular}{c@{ }c@{ }}
        \includegraphics[width=0.48\linewidth]{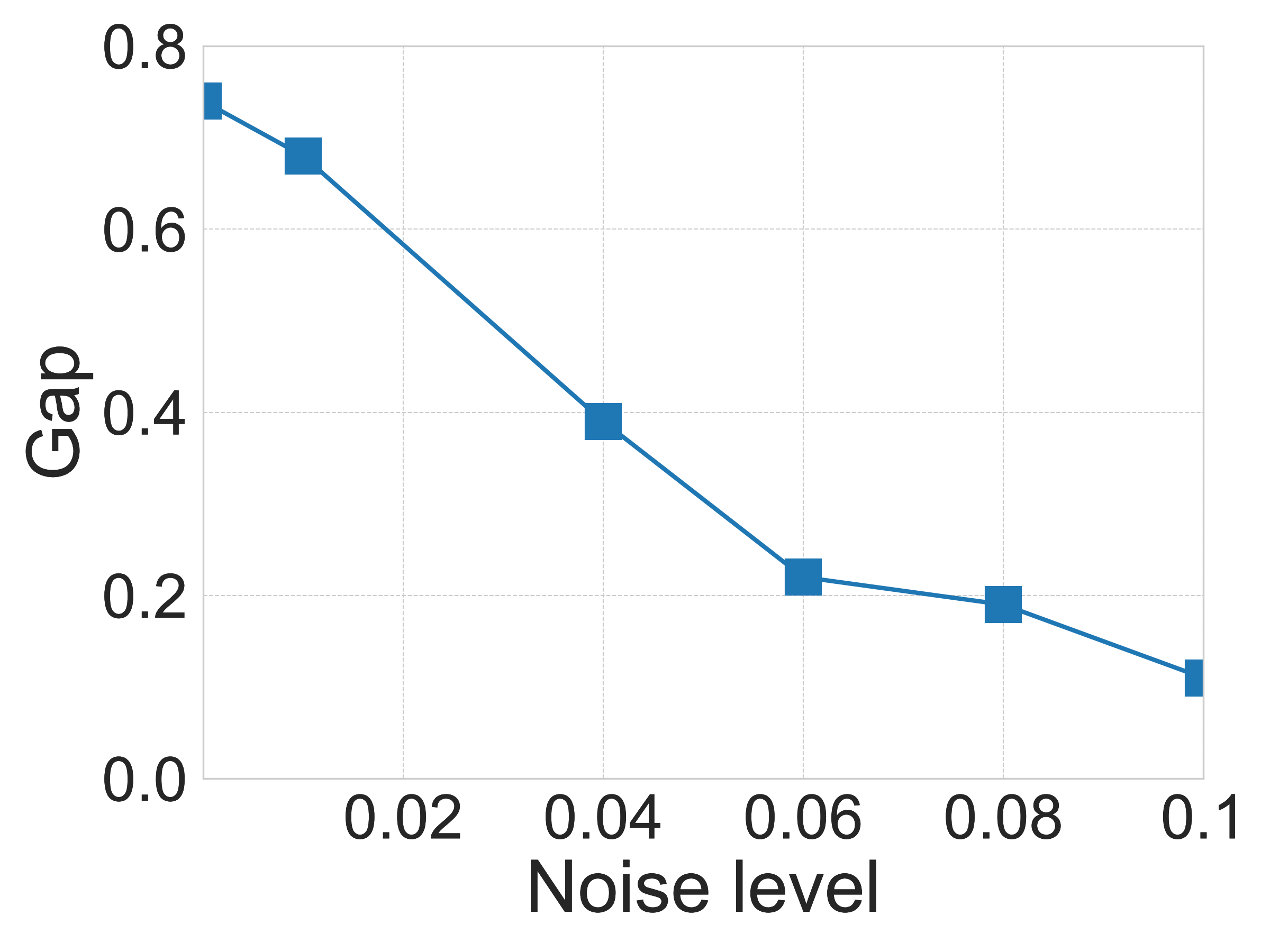}&
        \includegraphics[width=0.48\linewidth]{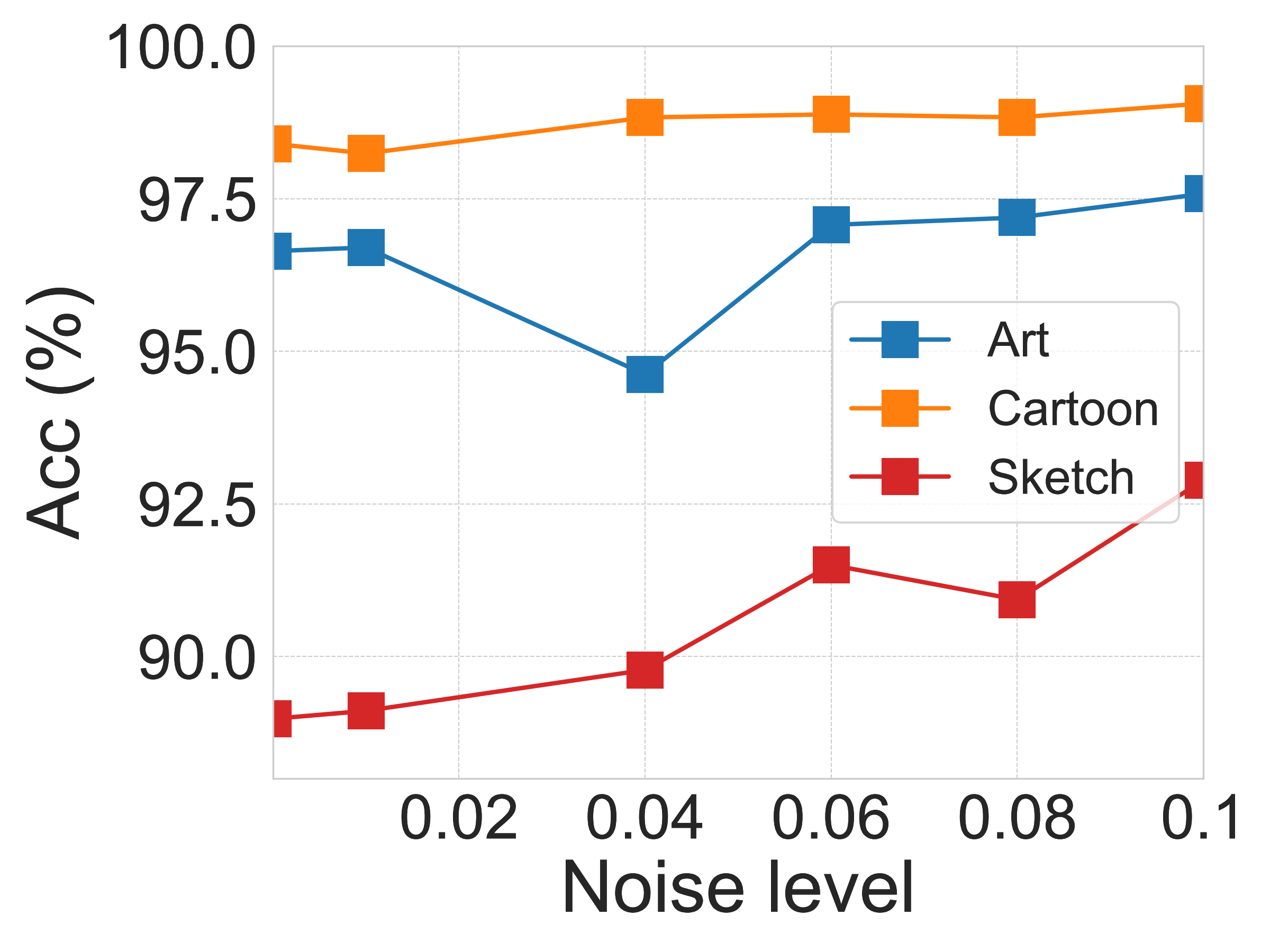}\\
    \end{tabular}
    \caption{\small From left to right. The effect of the noise level for modality gap and the effect of the noise level for performance on PACS.} 
    \label{fig_ab2}
\end{figure}
\noindent\textbf{Compare with generative models.} Similar to PODA \cite{poda}, we also compare our LDFS to CLIPStyler \cite{clipstyler}, the state-of-the-art text-guided zero-shot style transfer method that leverages CLIP with an additional generative model to generate novel domain images. In contrast, our method synthesizes novel domain image features. For both models, we sample 16 images per class from the photo domain and synthesize target domain features/images based on text descriptions (e.g., a sketch photo of a [CLS]). We also experiment with using image-specific descriptions identical to ours, denoted as CLIPStyler$\dagger$. They are then finetuned using CoOp \cite{coop}, with results shown in Table \ref{tab6}. We can see that CLIPStyler consistently underperforms our method and even underperforms the zero-shot CLIP in the art and cartoon domain. Upon examining CLIPStyler's synthetic images (Fig. \ref{figstyler}), we note its capability to generate target domain images but also its tendency to introduce unexpected artifacts, especially in cartoon domains (red boxes). Also, it sometimes fails at synthesizing the target class (e.g., the elephant in the sketch domain), negatively impacting the adaptation performance. Additionally, CLIPStyler$\dagger$ fails to surpass CLIPStyler despite using image-specific descriptions. We argue that CLIPStyler is designed for zero-shot style transfer and is worse at complex image-specific attribute generation.\\
\noindent\textbf{Analysis of the synthetic features quality on PACS.} We assess the quality of synthetic features in our LDFS against the SOTA text-guided feature synthesis method, LADS \cite{lads}. On the PACS dataset, given photo domain images, we synthesize novel features for other domains using both methods. Since neither method directly generates images, we employ nearest neighbor images from the respective target domains to verify synthetic feature quality. In Fig. \ref{fig4}, we display 8 randomly chosen samples from the photo domain and the corresponding synthetic target domain images. It shows LADS and LDFS successfully generate target domain features. Nevertheless, LADS often produces repetitive or similar features for the same category, as highlighted in the red boxes of Fig. \ref{fig4} (a). Meanwhile, LADS struggles to preserve class information, especially for those intricate samples during adaptation, as shown in green boxes of Fig. \ref{fig4} (a). In contrast to LADS, our LDFS can synthesize more diverse target domain features with superior quality while maintaining class consistency, as shown in Fig. \ref{fig4} (b). Furthermore, to quantitatively evaluate the quality of the synthesized features, we follow \cite{lads} in using the domain alignment (DA) score to evaluate the model's ability to correctly translate data to the target domain. We also use the class consistency (CC) score to evaluate how well the model preserves class information during synthesis. Further, we propose a feature diversity (DS) score to evaluate the diversity of the synthesized features. Specifically, the DA score is obtained by calculating the percentage of nearest neighbors belonging to the desired target domain. The CC score is obtained by calculating the percentage of synthetic features that belong to the original class. The DS score is obtained by one minus the percentage of repeated nearest neighbors within the same class. Table \ref{tab5} shows that LDFS achieves the highest domain alignment (DA), class consistency (CC), and diversity scores (DS), which further demonstrates the efficacy of our strategy. 
\begin{figure*}[t!]
    \centering
    \setlength{\tabcolsep}{3pt}  % Adjust this value to get desired spacing
    \renewcommand{\arraystretch}{0.8}
    \begin{tabular}{c@{\hspace{10pt}}c}
        \includegraphics[width=0.47\linewidth]{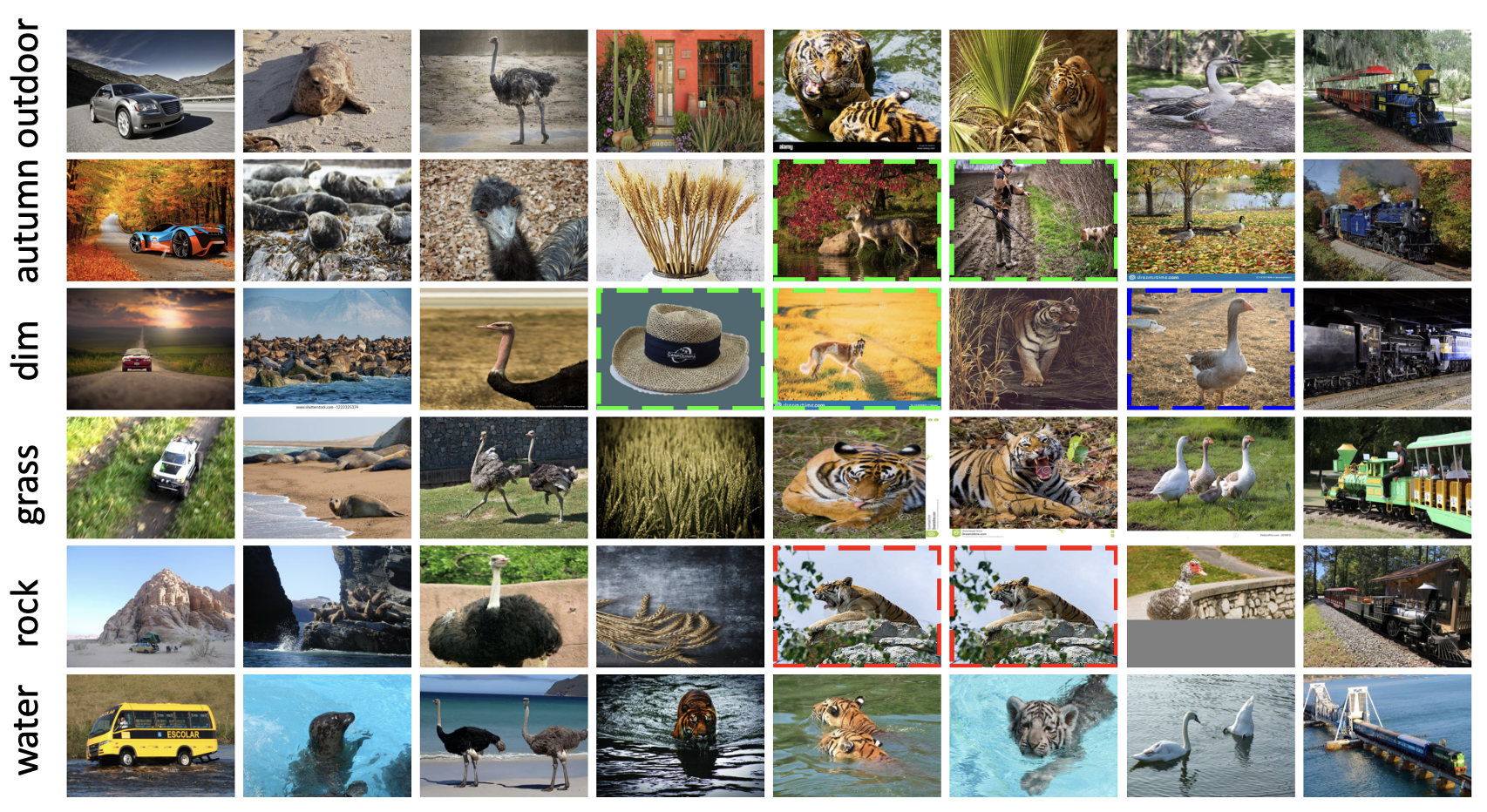} &
        \includegraphics[width=0.47\linewidth]{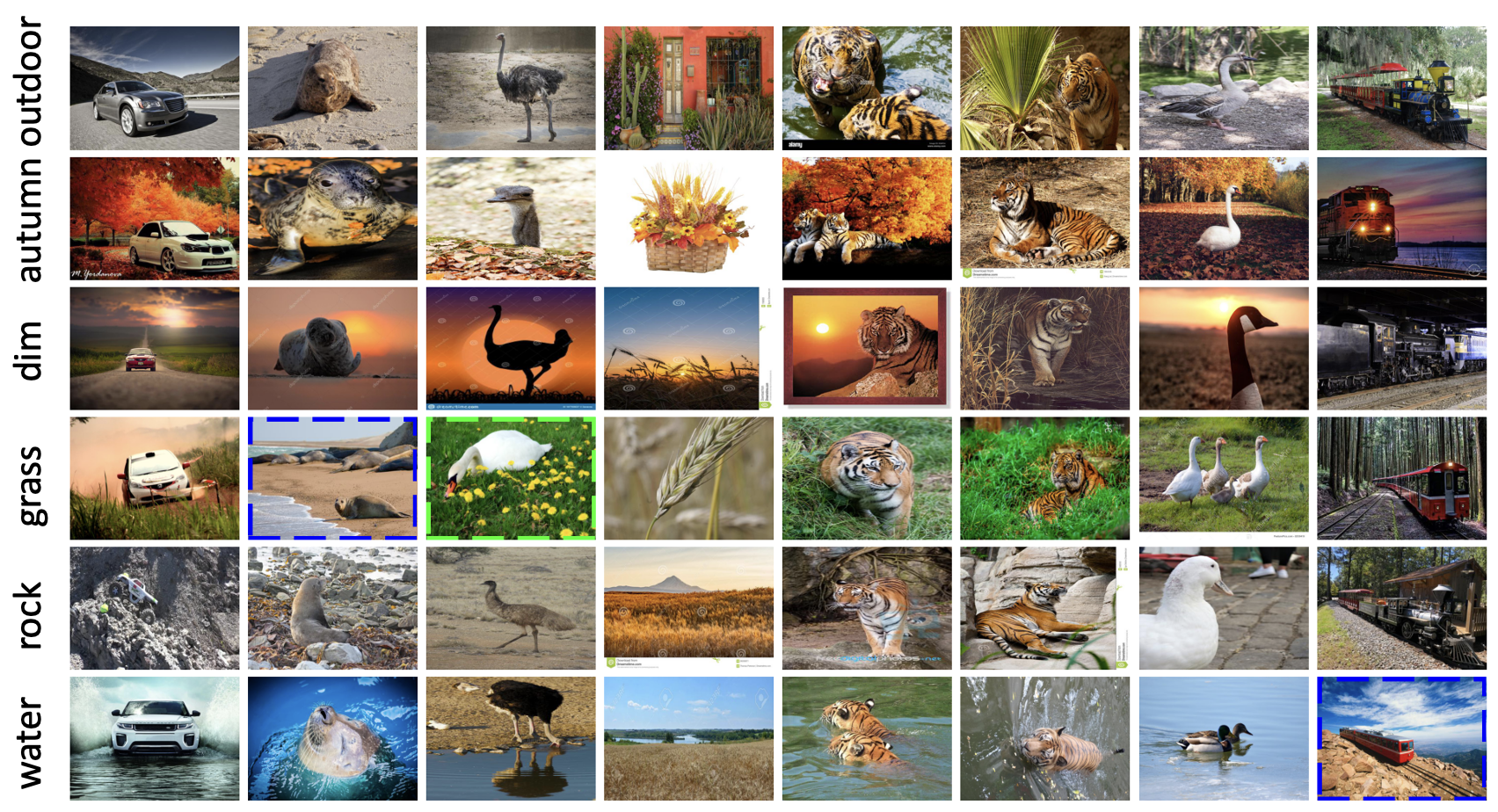} \\
        \footnotesize{(a) Synthesized feature using LADS \cite{lads}.} & \footnotesize{(b) Synthesized feature using LDFS (ours).} \\
    \end{tabular}

% \vspace{-2mm}
\caption{\small \textbf{Comparison of nearest neighbor images of synthesized target domain features between LADS and our LDFS on the NICO++ dataset: outdoor → autumn, dim, grass, rock, and water (zoom in for better visualisation).} The first row displays randomly sampled images from the source domain (outdoor). Subsequent rows showcase the nearest neighbors of synthetic features from both LADS and LDFS. Red boxes highlight issues with homogeneous synthetic features lacking diversity for both methods. Blue boxes indicate where both methods fail to generate target domain features. Green boxes show where both lose class information during adaptation, although LDFS preserves more class information than LADS.}

    \label{sup_f2}
\end{figure*}

\noindent\textbf{Analysis of the synthetic features quality on NICO++.}  Additionally, we compare the synthetic feature quality of our LDFS against LADS \cite{lads} on NICO++ dataset \cite{nico++}. Given outdoor domain images, we synthesize novel features for the autumn, dim, grass, rock, and water domains using both LDFS and LADS. In Fig. \ref{sup_f2}, we display 8 randomly chosen samples from the outdoor domain and corresponding target domains. Both LDFS and LADS are effective for synthesizing target domain features in most cases. However, LADS often produces repetitive features lacking diversity (red boxes in Fig. \ref{sup_f2} (a)) or wrong class features (green boxes in Fig. \ref{sup_f2} (b)). In contrast, our LDFS synthesizes more diverse, high-quality target domain features while maintaining class consistency. However, we observe both LDFS and LADS can sometimes generate wrong domain features (blue boxes in Fig. \ref{sup_f2}). This rarely occurred on the simpler PACS dataset, indicating the NICO++ domains are more complex and challenging. Moreover, Table \ref{sup_tab2} shows that our LDFS achieves the highest class consistency (CC) and diversity scores (DS), but a slightly lower domain alignment score (DA) compared to LADS \cite{lads}. Furthermore, it can be seen that the generative model CLIPStyler \cite{clipstyler} achieves a higher diversity score than LADS, but much lower domain alignment and class consistency scores.
\begin{table}[t!]

   \caption{\small Synthetic feature quality (\%) on NICO++ . }
   \centering
   \small
   \scalebox{0.9}{
    \setlength{\tabcolsep}{3.1mm}{
\begin{tabular}{lccc}
  \toprule[1pt]
 & LADS           & CLIPStyler       & LDFS           \\ \hline
DA score                        & \textbf{96.18 }         & 74.27          & 94.33              \\
% \hline
% \rowcolor{Gray2}
CC score   & 82.96 & 68.69 & \textbf{88.54}  \\                        
% \rowcolor{Gray2}
DS score                     & 76.86 & 85.81  & \textbf{92.45}   \\
 \midrule[1pt]
\end{tabular}}}
\label{sup_tab2}
\end{table}

\begin{table*}[t]
   \caption{Ablation on the effect of different text descriptions over all domains on NICO++ \cite{nico++}.}

   \centering
   \small
   \scalebox{0.85}{
    \setlength{\tabcolsep}{1.6mm}{
    \begin{tabular}{ccc}
    \toprule[1pt]
    Base Text Description
 & CoOp \cite{coop} & CoOp+LDFS \\
    \midrule
    ``a photo of a \{class name\} with \{domain name\} background.”
       & 81.95
 & 83.23 (\bluenum{+1.28})
\\
    ``a \{domain name\} style photo of a \{class name\}.”
 & 81.95 & 82.93 (\bluenum{+0.98})
\\
    ``a \{domain name\} background photo of a \{class name\}.”
 & 81.95 & 83.20 (\bluenum{+1.25})
 \\
    \bottomrule[1pt]
    \end{tabular}
    }
   }
\label{sup_tab3}
\end{table*}

\begin{table*}[t]
   \caption{Base text descriptions used in different datasets.}

   \centering
   \small
   \scalebox{0.8}{
    \setlength{\tabcolsep}{1.6mm}{
      \begin{tabular}{ccccc}
        \toprule[1pt]
                        & PACS                              & OfficeHome                         & DomainNet                           & NICO++   \\
        \midrule
        Source Prompts  & ``a real photo of a \{\}.''        & ``a real photo of a \{\}.''         & ``a realistic photo of a \{\}.''     & ``a photo of a \{\} with outdoor background.''      \\
        \hline
        {Target Prompts} 
                        & ``a art photo of a \{\}.''        & ``a clipart photo of a \{\}.''      & ``a sketch photo of a \{\}.''        & ``a photo of a \{\} with autumn background.''      \\
                        & ``a sketch photo of a \{\}.''      & ``a stock photo of a \{\}.''        & ``a painting photo of a \{\}.''      & ``a photo of a \{\} with dim background.''    \\
                        & ``a cartoon photo of a \{\}.''     & ``a sketch photo of a \{\}.''       & ``a clipart photo of a \{\}.''       & ``a photo of a \{\} with grass background.''   \\
                        & ``a painting photo of a \{\}.''    & ``a art photo of a \{\}.''          & ``a infograph photo of a \{\}.''     & ``a photo of a \{\} with rock background.''\\
                        & \text{-}                          & \text{-}                           & ``a quickdraw photo of a \{\}.''     & ``a photo of a \{\} with water background.''\\
        \bottomrule[1pt]
      \end{tabular}
    }
  }
  \label{tab2_sup}
\end{table*}

\subsection{Ablation Studies}
\label{sec_ab}
In this subsection, we focus on ablating the major components of our proposed framework. We provide additional ablations, including analyses of different text generation strategies, domain text descriptions, number of training images,  synthetic feature quality, etc. Also, when we refer to LDFS during ablation study, we are specifically referring to the LDFS + CoOp.\\
\noindent\textbf{Analysis of stochastic text feature augmentation.} We carry out experiments to investigate the effectiveness of our introduced stochastic text feature augmentation approach. Specifically, we analyze the effect of the noise level, denoted as $\gamma$, on the modality gap and subsequent performance. As depicted in the left panel of Fig.~\ref{fig_ab2}, there's a consistent decline in the modality gap as the noise level rises, demonstrating our method's ability to bridge this gap. Correspondingly, the right panel of Fig.~\ref{fig_ab2} illustrates that by closing the modality gap via our technique, LDFS performance is enhanced in target domain adaptation. For most noise levels, stochastic text feature augmentation can enhance LDFS performance compared to when there's no augmentation (i.e., noise level=0). The peak performance boost is observed at a noise level of 0.1 across all three domains. Additionally, referring to Table \ref{tab_ab}, it's evident that our method (i.e., LDFS), when using tuned $\gamma$, boosts LDFS performance without augmentation (i.e., LDFS w/o TA) by respective margins of 2.99\%, 1.51\%, and 0.31\% across PACS, OfficeHome, and NICO++ benchmarks. Compared to increasing the gap (LDFS w/TA+) and adding noise to image features (LDFS w/ IA), our text augmentation method (LDFS) performs the best.\\

\begin{figure}[!t]
    \centering
        \includegraphics[ width=0.95\linewidth]{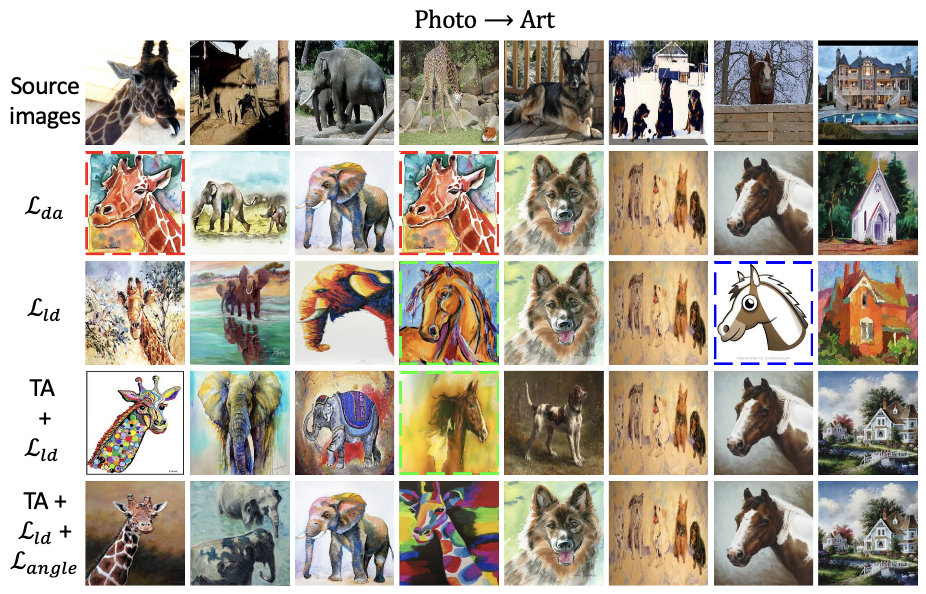} 
        % \caption{Illustration of the text-guided feature synthesis.} 
    \caption{\small \textbf{Nereast neighbor images of our LDFS when ablating different components on PACS.} $\mathcal{L}_{da}$ denotes global direction adaptation loss of LADS. Red boxes show the issue of homogenous synthetic features; blue boxes show the model fails to synthetic target domain features; and green boxes show the model fails to preserve class information during adaptation.}
    \label{fig7} 
\end{figure}

\begin{table}[t]
   \caption{Ablation studies on different datasets (\%).}

   \centering
   \small
   \scalebox{0.95}{
    \setlength{\tabcolsep}{1.6mm}{
\begin{tabular}{l|ccc}
\toprule[1pt]
Method  & PACS   & OfficeHome & NICO++\\ \hline
LDFS   & \textbf{96.51} & \textbf{82.96}   & \textbf{83.23}   \\
LDFS w TA+  & 94.68 & 81.49   & 82.82\\
LDFS w IA  &93.66  & 77.11   & 80.38\\
LDFS w/o TA  & 93.52 & 81.45   & 82.92     \\
LDFS w/ global  & 96.34 & 81.66   & 82.95  \\
LDFS w/o $\mathcal{L}_{pair}$   & 95.63& 80.87   & 83.04   \\
LDFS w/o TA, $\mathcal{L}_{pair}$   & 94.90& 80.49   & 82.65  \\
LDFS w/o TA, $\mathcal{L}_{pair}$,  $\mathcal{L}_{cc}$ 
& 92.89& 76.54   & 71.99  \\ 
 \midrule[1pt]
\end{tabular}}}
\label{tab_ab}
\end{table}

 \begin{figure}[t]
    \centering
    \setlength{\abovecaptionskip}{0.cm}
    \setlength{\belowcaptionskip}{-0.cm}
    \setlength{\tabcolsep}{1pt}  % <-- Adjust this value to get desired spacing
    \renewcommand{\arraystretch}{0.8}
    \begin{tabular}{c@{ }c@{ }}
            \vspace{-0.15cm}
        \includegraphics[width=0.48\linewidth]{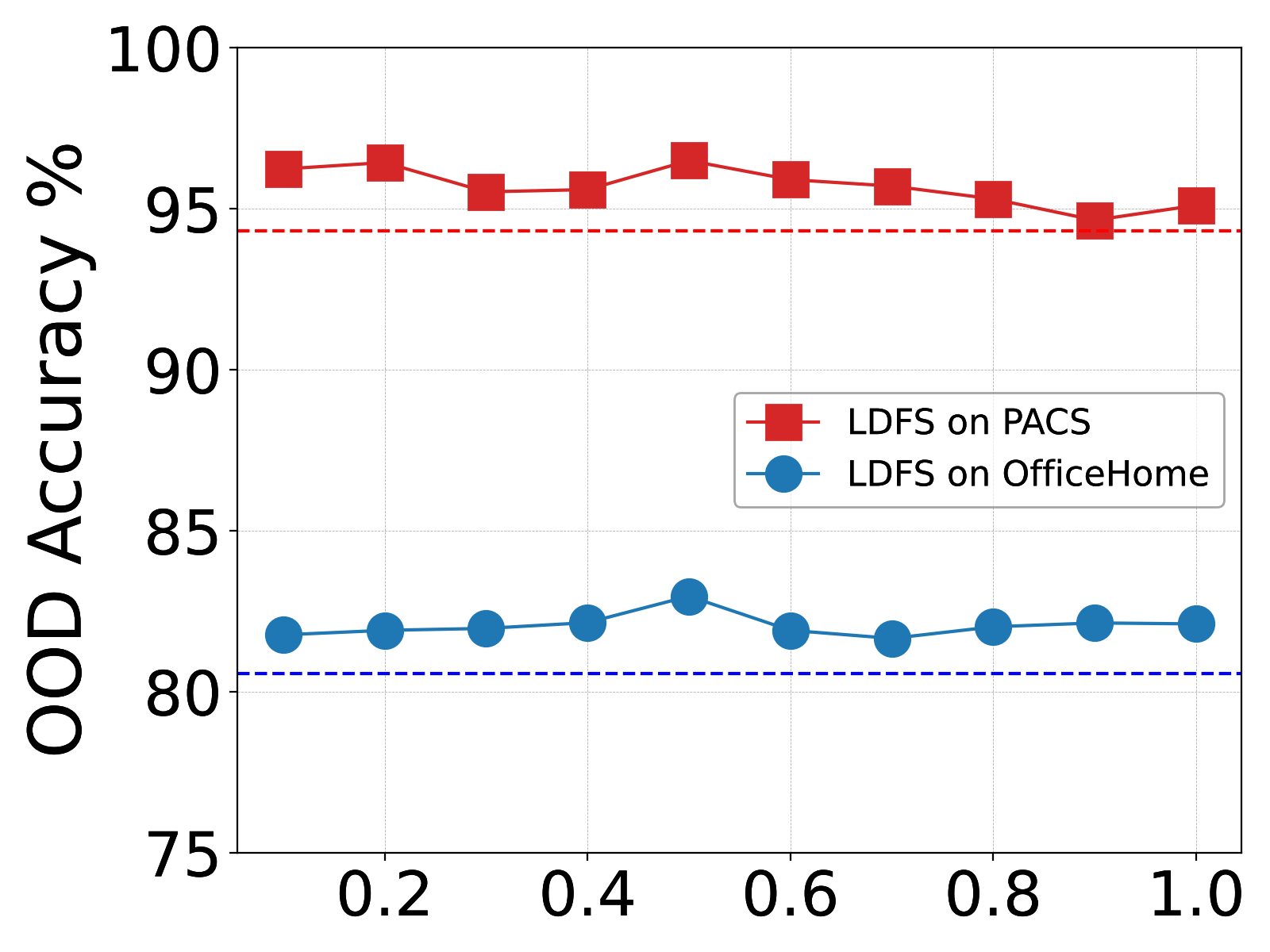}&
        \includegraphics[width=0.48\linewidth]{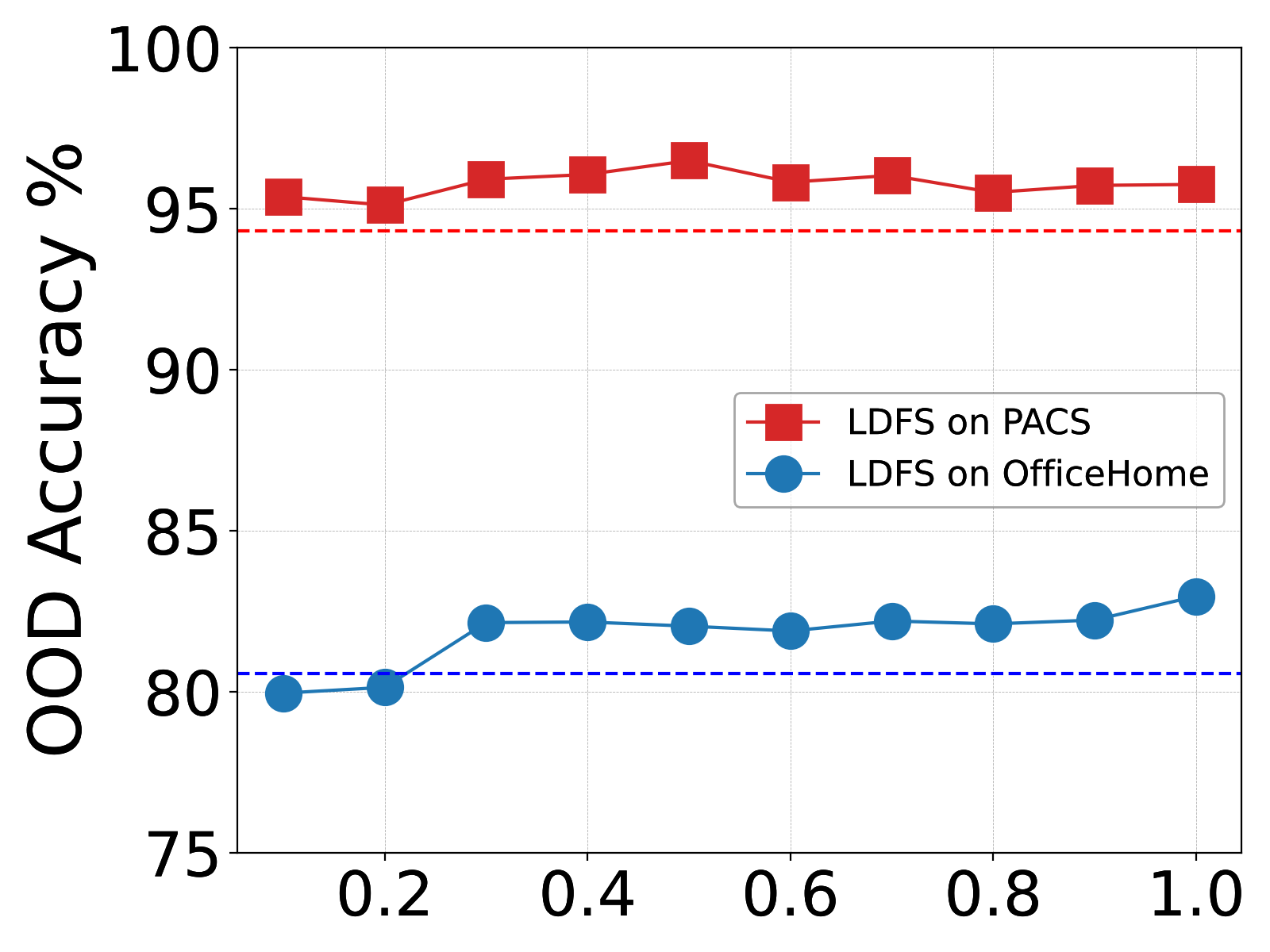}\\

            \hphantom{XX}\footnotesize{(a) $\alpha$}&\hphantom{XX}\footnotesize{(b) $\beta$}\\
    \end{tabular}
    \caption{\small Sensitivity analysis of the weighting hyperparameters $\alpha$ and $\beta$. The red and blue dash lines represent the performance of the model without using LDFS on PACS and OfficeHome.} 
    \label{fig_ab4}
\end{figure}

\begin{figure}[!t]
    \centering
        \includegraphics[ width=0.9\linewidth]{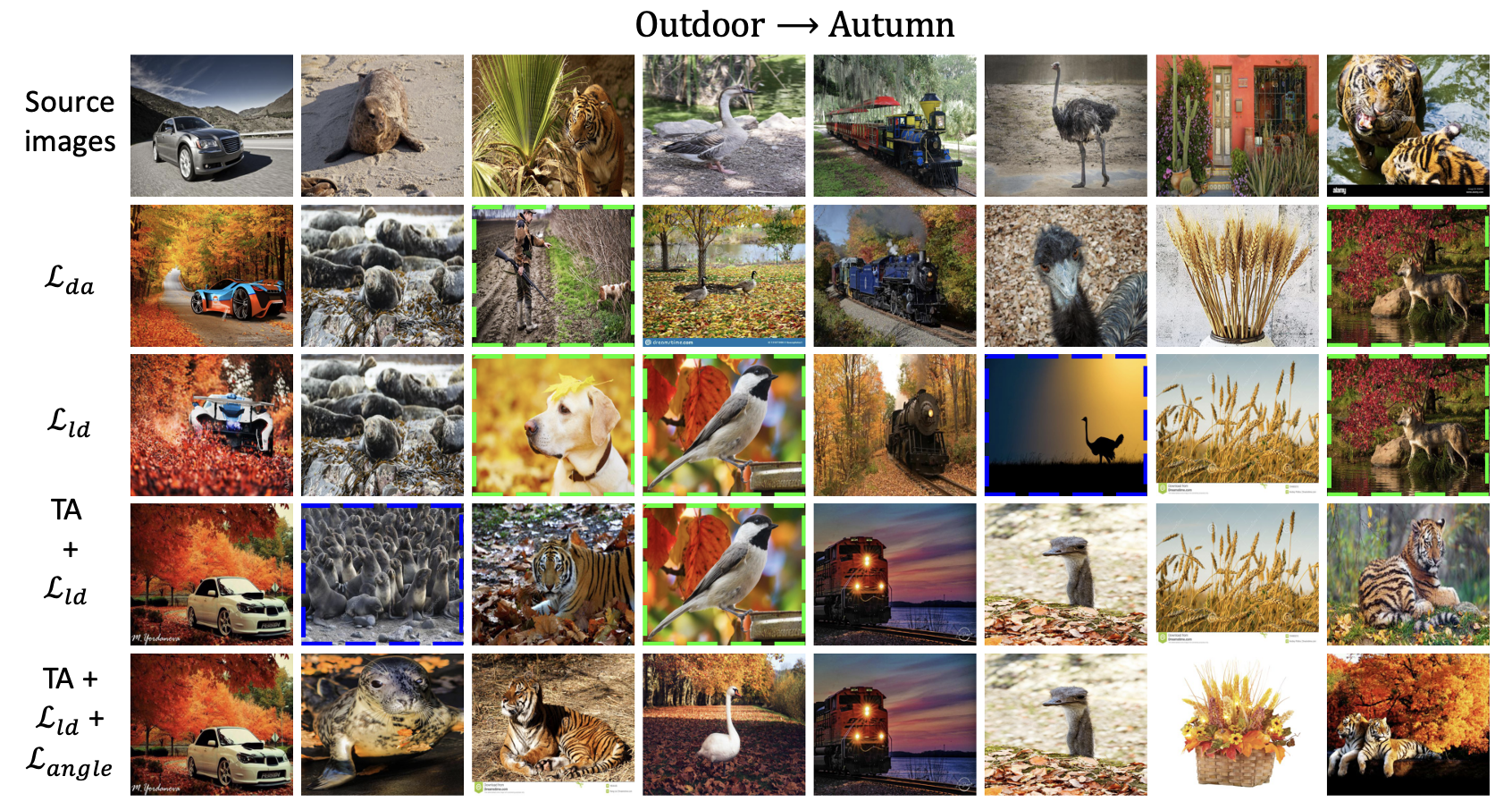} 
        % \caption{Illustration of the language-guided zero-shot domain adaptation.} 
     
          % \vspace{-3mm}
    \caption{\small \textbf{Nereast neighbor images of our LDFS when ablating different components on NICO++.} $\mathcal{L}_{da}$ denotes global direction adaptation loss of LADS. Red boxes show the issue of homogenous synthetic features; blue boxes show the model fails to synthetic target domain features; and green boxes show the model fails to preserve class information during adaptation.}
    \label{sup_f3} 
\end{figure}

\noindent\textbf{Ablation on the loss.}
In Table \ref{tab_ab}, we evaluate the effect of $\mathcal{L}_{ld}$, $\mathcal{L}_{pair}$, and $\mathcal{L}_{cc}$ used on our LDFS. LDFS w/ global denotes our baseline modified by replacing $\mathcal{L}_{ld}$ with the conventional global adaptation loss \cite{lads}. We can see that excluding the local adaptation losses $\mathcal{L}_{ld}$ or $\mathcal{L}_{pair}$ leads to performance degradation across all datasets. Also, removing $\mathcal{L}_{cc}$ results in significant degradation, indicating the importance of preserving category information during feature synthesis. Moreover, LDFS, when fully equipped with all components, outperforms all baselines across all datasets, showcasing the importance of all proposed components. From an alternative perspective on synthetic feature quality, we study the ablation results of translating features from the photo domain into the art domain, as illustrated in Fig.~\ref{fig7}. Red boxes highlight the issue of homogeneous feature synthesis, the blue box indicates models' failures in domain translation, and the green boxes show models struggle to preserve class information. $\mathcal{L}_{da}$ indicates baseline with global adaptation loss. We observe that replacing $\mathcal{L}_{da}$ with our local adaptation loss $\mathcal{L}_{ld}$ can address the diversity issue, yet result in the synthesis of incorrect class or domain features. Combining all components together leads to the best quality of synthetic features. Similar observations can also be found on NICO++ dataset, as shown in Fig. \ref{sup_f3}. \\
\begin{table}[t]
   \caption{Ablation on Different descriptions(\%).}
   \centering
   \small
   \scalebox{1}{
    \setlength{\tabcolsep}{3mm}{
\begin{tabular}{l|ccc}
\toprule[1pt]
Method  & PACS   & OfficeHome & NICO++\\ \hline
LDFS   & \textbf{96.51} & \textbf{82.96}   & \textbf{83.23}   \\
LDFS\_en   & 95.97 & 82.40   & 82.19   \\

LDFS w/ global  & 96.34 & 81.66   & 82.95  \\
WaffleCLIP \cite{waffleclip}   & 95.37& 80.34  & 81.19  \\
CLIP (ZS)&94.77&79.27&80.04\\
\midrule[1pt]
\end{tabular}}}
\label{sup_tab5}
\end{table}
\noindent\textbf{Analysis of Instance-conditional Loss.} We compared our loss variant using averaged random descriptions, inspired by WaffleCLIP \cite{waffleclip}, denoted as $LDFS\_en$. However, we found no significant improvement over using a single description. Additionally, our local adaptation strategy substantially outperformed the global adaptation approach (i.e., LDFS w/ global). Our observations indicate that more semantically relevant and diverse descriptions lead to improved feature synthesis and performance gains. In contrast, simply adding or averaging stochastic directions does not yield comparable improvements. These findings underscore the importance of quality and diversity in descriptions rather than mere quantity or randomness.\\
{\noindent\textbf{The Effect of Text Description.} To explore if our method relies on users' verbalization skills, we ablate different text descriptions in Table \ref{sup_tab3}. It shows consistent efficacy despite variations in description format. It can also be seen that all prompts improve over the CoOp \cite{coop} baseline, with ''a photo of a {class name} with {domain name} background'' achieving the best result. Moreover, Table \ref{tab2_sup} displays straightforward, uniform text descriptions used across all our datasets in our paper. The simplicity of these descriptions demonstrates that our method is easily applicable and user-friendly.}\\
\noindent\textbf{Sensitivity analysis of $\alpha$ and $\beta$.}
 The weighting hyperparameters $\alpha$ and $\beta$ in Eq. \ref{eq4} play crucial roles in maintaining category consistency and pairwise relationship strength. From Fig.~\ref{fig_ab4} (a), it's evident that an $\alpha$ value of 0.5 leads to optimal performance on both PACS and OfficeHome. Additionally, Fig.~\ref{fig_ab4} (b) shows that increasing the contribution of $\mathcal{L}_{pair}$ enhances the performance of LDFS on these datasets, which ensures adapted features lie to the CLIP sphere, thus improving feature quality. We find optimal results are generally achieved with $\beta$ values of 0.5 or 1.\\
  \begin{figure}[t!]
    \centering
    \setlength{\abovecaptionskip}{0.cm}
    \setlength{\belowcaptionskip}{-0.cm}
    \setlength{\tabcolsep}{1pt}  % <-- Adjust this value to get desired spacing
    \renewcommand{\arraystretch}{0.5}
    \begin{tabular}{c@{ }}
        \includegraphics[width=0.5\linewidth]{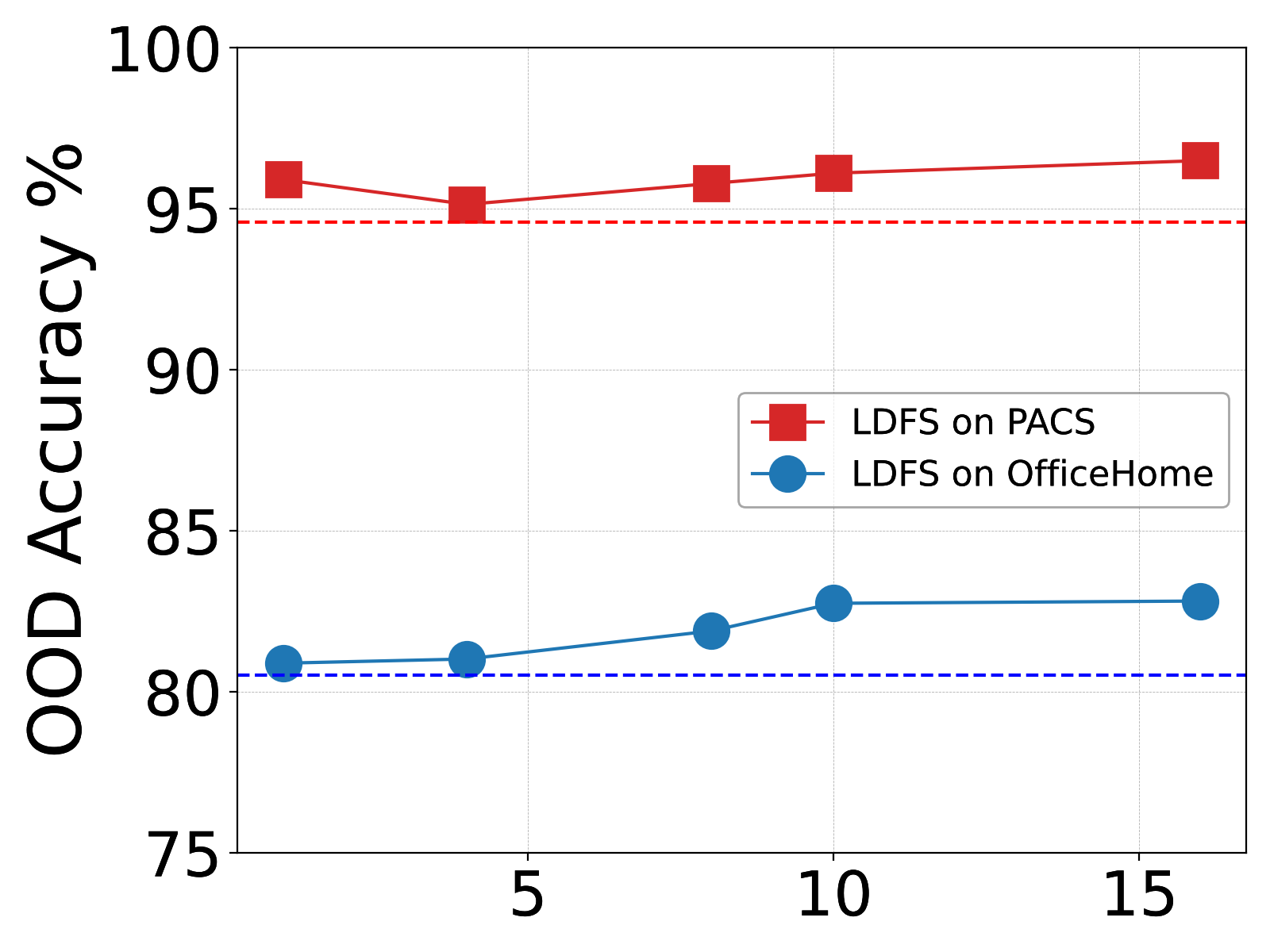}
          \\
    \end{tabular}
    \caption{\small Effect of number of shots on PACS and OfficeHome. The red and blue dash lines represent the performance of the model without using LDFS on PACS and OfficeHome.} 
    \label{sup_fig3}
\end{figure}
\noindent\textbf{Effect of Number of Shots.} In Fig. \ref{sup_fig3}, we plot the average performance of our LDFS framework using different few-shot settings (1, 4, 8, 10, and 16 shots per class) on the PACS \cite{pacs} and OfficeHome \cite{oh} datasets. As can be seen, the performance of LDFS overall increases as the number of shots increases. All settings with LDFS outperform the CoOp baseline without LDFS, even when using just 1 shot per class.\\
\noindent\textbf{Scope of Applications.} Limitation exists in our approach and all text-guided image/feature synthesis methods, as efficacy mainly relies on domains that can be described linguistically. Still, the task is practical and common since domains like weather, backgrounds, time of day, and colors frequently appear in real-world scenarios and are easy to describe. Our method fits most domain adaptation datasets. Moreover, our approach does not require target domain data, which is a significant advantage over its limitation.
\section{Conclusion}
In this paper, we present LDFS, a text-guided feature augmentation method. Through careful design, our proposed framework successfully synthesizes diverse visual features from novel domains with high quality. Extensive experiments demonstrate the effectiveness of LDFS in improving CLIP's generalization ability on unseen domains without the need for additional data collection from those domains.

% We also ablate different text descriptions in Table 4 in the Appendix, showing our method is consistently effective. Further, Table 2 in the Appendix shows the text descriptions used in all datasets in our paper are consistent and very straightforward formats. Further, our method does not need target domain data, a significant advantage over limitations.

% \textit{Scope of application:} Although LDFS enables domain adaptation through language without target domain data, its efficacy is mainly in domains that can be clearly described by language (\eg~different colors, backgrounds, weathers). Extending LGDA tasks to domains that are hard to describe (\eg~medical image analysis, impacted by domains like different hospital or camera setups) is a crucial future step. 

\bibliographystyle{splncs04}
\bibliography{main}

\vfill

\end{document}